\documentclass{article}

\PassOptionsToPackage{numbers,square}{natbib}
\usepackage[preprint]{neurips_2026}

\usepackage[utf8]{inputenc}
\usepackage[T1]{fontenc}
\usepackage{hyperref}
\usepackage{url}
\usepackage{booktabs}
\usepackage{amsfonts}
\usepackage{nicefrac}
\usepackage{microtype}
\usepackage{xcolor}
\usepackage{amsmath}
\usepackage{amssymb}
\usepackage{graphicx}
\usepackage{siunitx}
\title{ScaleMAP: Preserving Local Density and Neighborhood Structure in Low-Dimensional Embeddings}

\author{
Rajas Poorna\\
School of Chemical and Biomolecular Engineering\\
Georgia Institute of Technology\\
\texttt{rajasp@gatech.edu}
\and
Marcus T. Cicerone\\
School of Chemistry and Biochemistry\\
Georgia Institute of Technology\\
\texttt{cicerone@gatech.edu}
}

\begin{document}

\maketitle

\begin{abstract}
Nonlinear dimensionality-reduction methods such as UMAP and PaCMAP adaptively normalize local distances during graph construction, erasing neighborhood scale from the data. This distorts more than relative cluster sizes: sparse structures like bridges between transitioning cell types and narrow spectral spikes in hyperspectral images can be suppressed or lost entirely. DensMAP adds a density penalty to correct this, but this penalty competes with UMAP's attraction--repulsion forces, scattering points far from their neighborhoods. ScaleMAP takes a different approach: each pairwise embedding displacement is divided by the geometric mean of the two endpoints' original-space local radii, re-injecting scale information as a change of variables rather than as a competing objective. Across standard benchmarks and scientific datasets from transcriptomics, hyperspectral imaging, and flow cytometry, ScaleMAP matches DensMAP on density preservation while maintaining UMAP-level neighborhood preservation. In transcriptomic data, it recovers sparse bridges between cell populations that UMAP collapses; in flow cytometry, it faithfully represents density structure across 17 orders of magnitude. The same principle applied to PaCMAP yields consistently improved density preservation, suggesting the approach generalizes beyond UMAP.
\end{abstract}

\section{Introduction}
Low-dimensional embeddings are often the first step in exploratory analysis of large high-dimensional datasets. Nonlinear dimensionality-reduction methods---UMAP \citep{mcinnes2018umap}, t-SNE \cite{maaten2008visualizing}, PaCMAP \cite{wang2021understanding}---build an approximate $k$-nearest-neighbor graph in high-dimensional space and place points in two dimensions so as to represent that graph. UMAP in particular has become a default tool for exploratory analysis across the natural sciences. These methods normalize each point's distances to its neighbors by an adaptive bandwidth, which makes them robust to varying local density. Intuitively, one might imagine that neighborhood-preserving methods would preserve both neighborhood adjacency and scale, but, despite estimating the relevant scale during graph construction, they generally do not incorporate them into the embedding objective. Consequently, two regions with very different densities become indistinguishable to the embedding objective---and the consequences go beyond relative cluster sizing. Sparse structures such as bridges between transitioning cell types, narrow spectral spikes in hyperspectral images, and local densities spanning many orders of magnitude, such as in flow cytometry, can all be suppressed or lost.

DensMAP \cite{narayan2021assessing} addresses this by adding a penalty term to the UMAP loss that discourages mismatches between local density in the original and embedding spaces. However, this penalty competes with UMAP's attraction--repulsion forces, and DensMAP scatters a fraction of points far from their true neighbors, making sparse regions of the embedding difficult to interpret.

If UMAP represents the shape correctly, but not the scale, this suggests an intuitive approach to resolve this issue: rescale the space locally to match the scale in the original space. We can accomplish this by treating the distances between embedding points as non-uniform. ScaleMAP implements this by performing a \emph{change of variables}: each pairwise embedding displacement $\mathbf{y}_i - \mathbf{y}_j$ is divided by the geometric mean $\sqrt{r_i r_j}$ of the two endpoints' original-space local radii, leaving the UMAP objective otherwise unchanged. Because the rescaling applies uniformly to all forces, the local equilibrium that gives UMAP its neighborhood preservation is maintained while original-scale information is reintroduced.

\textbf{Contributions.}
\begin{enumerate}
\item We introduce ScaleMAP, a modification of UMAP that injects local-scale information through a change of variables rather than an additional loss term. The same principle transfers to PaCMAP.
\item Across standard benchmarks and three scientific data modalities, ScaleMAP attains UMAP-level neighborhood preservation while broadly matching or exceeding DensMAP on density preservation, with substantially fewer scattered points.
\item ScaleMAP recovers structures that UMAP collapses---transcriptomic bridges consistent with known developmental relationships, sparse spectral spikes in hyperspectral imaging---and preserves density faithfully across 17 orders of magnitude in flow-cytometry data.
\end{enumerate}

\section{Method}

\subsection{UMAP as attractive and repulsive forces}

UMAP \cite{mcinnes2018umap} constructs a weighted kNN graph in the original space. For each point $\mathbf{x}_i$ with $k$ nearest neighbors, define the directed membership weight
\begin{equation}
\label{eq:umap-weight}
\mu_{ij} = \exp\!\left( -\frac{d(\mathbf{x}_i, \mathbf{x}_j) - \rho_i}{\sigma_i} \right),
\end{equation}
where $\rho_i$ is the distance from $\mathbf{x}_i$ to its nearest neighbor and $\sigma_i$ is an adaptive bandwidth. Points that are close relative to $\sigma_i$ receive weight near 1; points that are far receive weight near 0. The bandwidth $\sigma_i$ is chosen for each point so that the total weight to its $k$ neighbors equals $\log_2 k$, ensuring that every point has a comparable number of effective neighbors regardless of local density. These directed weights are symmetrized to $v_{ij} = \mu_{ij} + \mu_{ji} - \mu_{ij}\mu_{ji}$.

The embedding $\mathbf{y}_i \in \mathbb{R}^2$ is initialized by spectral decomposition of the graph Laplacian \cite{belkin2003laplacian} and then refined by stochastic gradient descent on a cross-entropy loss \cite{mcinnes2018umap}, which decomposes into per-edge attractive and repulsive update steps. Writing $d_{ij}^2 = \|\mathbf{y}_i - \mathbf{y}_j\|^2$ for the embedding-space squared distance:

\textbf{Attractive step.} For a graph edge $(i,j)$ sampled with probability proportional to $v_{ij}$, move $\mathbf{y}_i$ toward $\mathbf{y}_j$:
\begin{equation}
\label{eq:umap-attr}
\mathbf{y}_i \;\leftarrow\; \mathbf{y}_i \;-\; \alpha \cdot \frac{2ab\,(d_{ij}^2)^{b-1}}{1 + a\,(d_{ij}^2)^b} \;(\mathbf{y}_i - \mathbf{y}_j).
\end{equation}

\textbf{Repulsive step.} For a randomly sampled non-neighbor $k$, push $\mathbf{y}_i$ away from $\mathbf{y}_k$:
\begin{equation}
\label{eq:umap-rep}
\mathbf{y}_i \;\leftarrow\; \mathbf{y}_i \;+\; \alpha \cdot \frac{2b}{(d_{ik}^2 + \epsilon)(1 + a\,(d_{ik}^2)^b)} \;(\mathbf{y}_i - \mathbf{y}_k),
\end{equation}
where $\alpha$ is the learning rate, $\epsilon$ is a small constant, and $a,b$ are fixed by calibration to UMAP's \texttt{min\_dist} parameter. The adaptive bandwidth $\sigma_i$ makes these updates robust to varying local density, but it also discards information about the absolute scale of neighborhoods.

\subsection{Local radius}

DensMAP \cite{narayan2021assessing} introduced the \emph{local radius}, a measure of original-space neighborhood scale:
\begin{equation}
\label{eq:local-radius}
r_i^2 = \frac{\sum_j v_{ij} \, d(\mathbf{x}_i, \mathbf{x}_j)^2}{\sum_j v_{ij}}.
\end{equation}
This is a weighted-RMS distance from $\mathbf{x}_i$ to its UMAP-graph neighbors, reusing the existing weights $v_{ij}$ without introducing additional hyperparameters. Computation is $O(n)$ given the kNN graph.

\subsection{DensMAP: a competing objective}

DensMAP maximizes the Pearson correlation between $\log r_o$ (original-space local radius) and $\log r_e$ (embedding-space local radius) by adding a penalty to the UMAP loss: $\mathcal{L}_{\text{DensMAP}} = \mathcal{L}_{\text{UMAP}} - \lambda\,\text{Corr}(\log r_o,\,\log r_e)$. In terms of the per-edge updates, this adds a density gradient to the attractive step while leaving repulsion unchanged:
\begin{align}
\label{eq:densmap-attr}
\textbf{Attractive:}\quad \mathbf{y}_i \;&\leftarrow\; \mathbf{y}_i \;+\; \Delta \mathbf{y}_i^{\,\text{attr}} \;+\; \lambda\,\nabla_{\mathbf{y}_i}\text{Corr}(\log r_e,\,\log r_o), \\
\label{eq:densmap-rep}
\textbf{Repulsive:}\quad \mathbf{y}_i \;&\leftarrow\; \mathbf{y}_i \;+\; \Delta \mathbf{y}_i^{\,\text{rep}},
\end{align}
where the gradient of the correlation involves partial derivatives of the embedding local radius with respect to $d_{ij}^2$ (see \cite{narayan2021assessing} for the full expression). Because the density gradient can oppose the attractive force while the repulsive force is unmodified, DensMAP can scatter points away from their neighborhoods---in our experiments, the misplaced-point fraction is sometimes up to an order of magnitude larger than UMAP's. \citet{wang2021understanding} demonstrated in the context of UMAP's sensitivity to the embedding initialization that UMAP's attractive force drops off quickly once the point is far from its local neighborhood, which can cause disconnected points to never reach their neighborhood again.

\subsection{ScaleMAP: a change of variables}

Rather than adding a competing objective, ScaleMAP modifies the distance that enters the existing UMAP forces. Define the \emph{rescaled squared distance}
\begin{equation}
\label{eq:rescaled-dist}
\tilde{d}_{ij}^2 = \frac{d_{ij}^2}{r_i\,r_j} = \frac{\|\mathbf{y}_i - \mathbf{y}_j\|^2}{r_i\,r_j}.
\end{equation}
Substituting $\tilde{d}_{ij}^2$ for $d_{ij}^2$ in the cross-entropy loss and differentiating yields the ScaleMAP update rules. The chain rule introduces a factor of $1/(r_i r_j)$ in the displacement direction:

\textbf{Attractive step.}
\begin{equation}
\label{eq:scalemap-attr}
\mathbf{y}_i \;\leftarrow\; \mathbf{y}_i \;-\; \alpha \cdot \frac{2ab\,(\tilde{d}_{ij}^2)^{b-1}}{1 + a\,(\tilde{d}_{ij}^2)^b} \;\frac{\mathbf{y}_i - \mathbf{y}_j}{r_i\,r_j}.
\end{equation}

\textbf{Repulsive step.}
\begin{equation}
\label{eq:scalemap-rep}
\mathbf{y}_i \;\leftarrow\; \mathbf{y}_i \;+\; \alpha \cdot \frac{2b}{(\tilde{d}_{ik}^2 + \epsilon)(1 + a\,(\tilde{d}_{ik}^2)^b)} \;\frac{\mathbf{y}_i - \mathbf{y}_k}{r_i\,r_k}.
\end{equation}

The coefficient functions are identical to UMAP's (\ref{eq:umap-attr}--\ref{eq:umap-rep}), evaluated at $\tilde{d}^2$ instead of $d^2$. No new loss term is introduced. Because the rescaling enters both the attractive and repulsive forces symmetrically, the local equilibrium that gives UMAP its neighborhood preservation is maintained. Points in sparse regions (large $r_i, r_j$) experience weaker forces per unit Euclidean displacement, so the embedding spreads them out; points in dense regions are kept tightly packed. We show in Section~\ref{sec:ablations} that replacing the geometric mean $\sqrt{r_i r_j}$ with either endpoint alone destroys the embedding.

\subsection{Computational overhead and default settings}

The local radius computed in the original space has a distance scale determined by the data, while the UMAP embedding objective has an independent distance scale. We thus normalize the local radii before substituting them in (\ref{eq:rescaled-dist}):

\begin{equation}
\label{eq:scalemap-percentile}
r_{\mathrm{normalized}}(i)
=
\frac{r_{\mathrm{original}}(i)}
{P_{\eta}\!\left(\{r_{\mathrm{original}}(j)\}_{j=1}^{n}\right)}.
\end{equation}

Here, $n$ denotes the total number of points, and $P_\eta$ is the normalizing factor, the $\eta^{th}$ percentile of the local radii. We use the 95$^{th}$ percentile, $P_{95}$, as a default for all experiments; we examine sensitivity to this parameter in Appendix \ref{sec:percentile_dependence}. :

We start with the same spectral initialization as UMAP and apply this modified update function at every iteration, unlike DensMAP, which only applies the modified update function on a fraction of the iterations.

ScaleMAP adds $O(n)$ overhead to UMAP, identical to DensMAP: only the per-point local radii must be precomputed. We parallelized this step, which can otherwise be a bottleneck on large datasets. However, in comparisons between methods, we do not modify DensMAP's implementation. We use 800 epochs by default for ScaleMAP (versus 200 for UMAP, 400 for DensMAP). Although ScaleMAP gives qualitatively nearly identical embeddings at both 200 and 800 epochs for most datasets, the fraction of disconnected points is larger than UMAP when both are set at 200 epochs. Using more epochs results in a modest (6-33\%) reduction in the fraction of disconnected points while still retaining practical runtimes. See Appendix \ref{sec:extended_benchmarks} for comparisons.

\subsection{Extension to PaCMAP}

PaCMAP \cite{wang2021understanding} computes a per-point distance scale $\sigma_i$ (the mean distance to the 4th--6th nearest neighbors) for neighbor selection; pairwise distances are divided by $\sigma_i \sigma_j$ when constructing the graph, though these scaled distances are not used during optimization. We repurpose $\sigma_i$ in the embedding by replacing $r_i r_j$ with $\sigma_i \sigma_j$ in the rescaled distance (\ref{eq:rescaled-dist}) and applying the analogous change of variables to PaCMAP's attractive and repulsive updates. We call this Scale-PaCMAP and treat it as a portability check rather than a co-equal method. We also use twice as many iterations as PaCMAP's default at each stage for the same reason as ScaleMAP.

\section{Experimental setup}

\paragraph{Datasets.}
We evaluate on standard benchmarks---MNIST\cite{lecunmnist}, Fashion-MNIST \cite{xiao2017fashion}, COIL-20 \cite{nene1996coil20}, and Mammoth \cite{smithsonianmammoth}---and on three scientific datasets: 264{,}824 Tabula Sapiens immune cells \cite{tabulasapiens2022} represented by the provided 50-dimensional SCVI latent embedding \cite{lopez2018deep}; a $610\times610$ BCARS \cite{camp2014highspeed} hyperspectral image of a \emph{C.~elegans} gonad with 650 spectral channels \cite{poorna2023toward}; and human bone-marrow flow-cytometry data with 8 raw fluorescence marker channels \cite{qiu2011spade}. Two synthetic diagnostics, XOI and Bridge, are described in Section~\ref{sec:synthetic}.

\paragraph{Metrics.}
For density preservation, we compute local radii in the original and embedding spaces and report the $R^2$ of a linear fit between $\log r_e$ and $\log r_o$, reflecting the relationship between hypervolumes across the two spaces. For neighborhood preservation, we use \emph{kNN recall} at $k{=}15$, calculated as the average fraction of the $k$ original space nearest neighbors that are retained as nearest neighbors in the embedding. We measure extreme errors in neighborhood preservation by counting \emph{disconnected} points, which are points such that none of the point's $k{=}100$ nearest neighbors in the embedding are among its $k{=}100$ nearest neighbors in the original space. This is a conservative lower bound on misplaced points: it does not detect pairs of neighbors that are jointly ejected from their true neighborhood, which we do observe in DensMAP embeddings. For \emph{class mixing}, we report the fraction of originally label-pure points, defined as points all of whose $k{=}15$ nearest neighbors share their label, which become label-impure in the embedding, i.e., do not all share their label.

\paragraph{Baselines and protocol.}
We compare ScaleMAP with UMAP and DensMAP using their Python implementations and default hyperparameters except where noted. Each method was run five times using distinct random seeds. Because UMAP-family implementations use multithreaded stochastic optimization, these seeds do not make runs bitwise deterministic; variation across runs reflects both seeded randomness and nondeterminism from parallel execution. Main tables report means, with error bars in Appendix \ref{sec:extended_benchmarks}.

\section{Results}

\begin{figure}[htbp]
\centering
\includegraphics[width=1.0\textwidth]{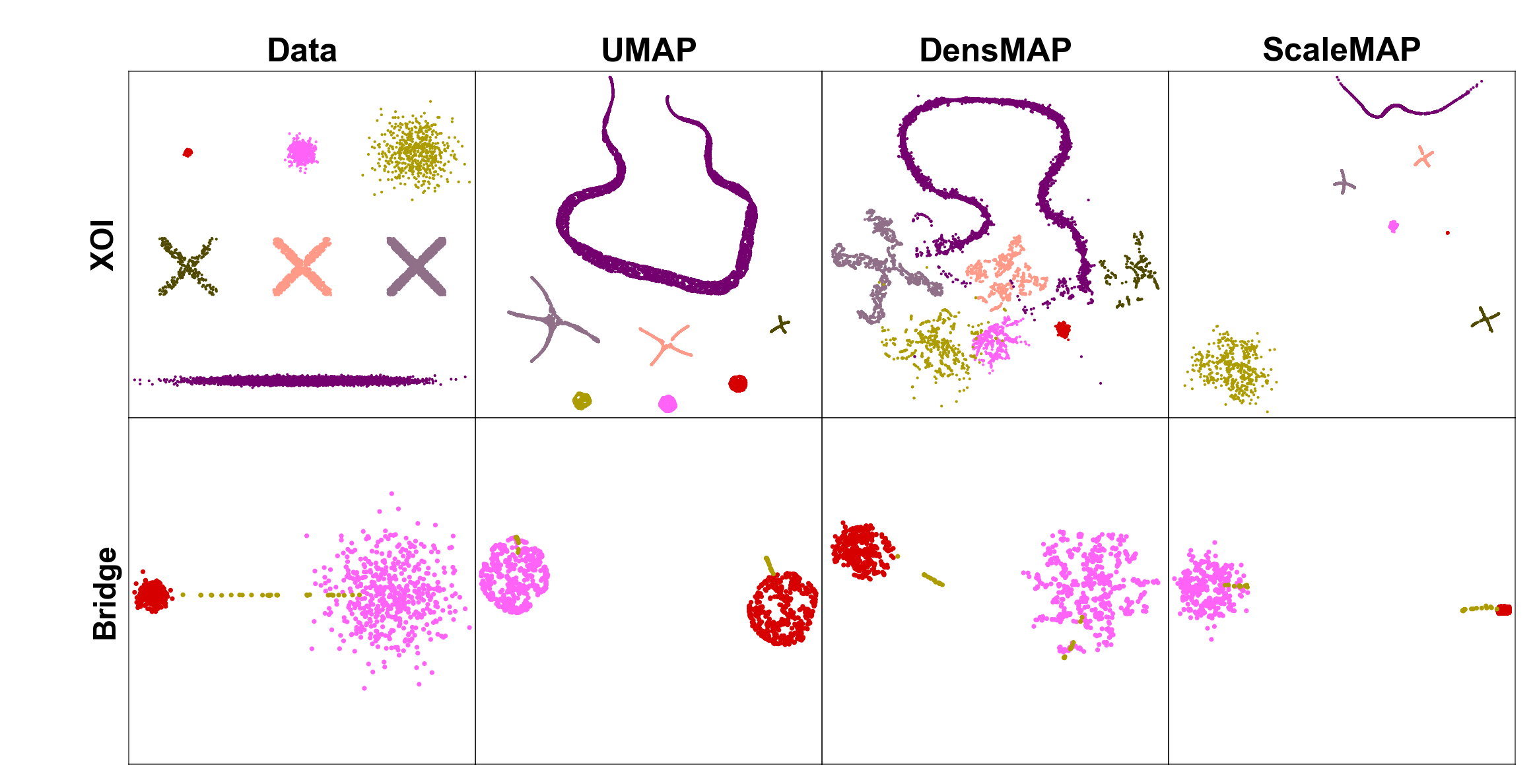}
\caption{\textbf{Synthetic diagnostics.} Comparison of UMAP, DensMAP, and ScaleMAP on two datasets designed to isolate density-preservation failures. \emph{Top (XOI):} three X shapes of equal spatial extent but increasing point count, three Gaussians of equal count but increasing standard deviation, and one high-aspect-ratio Gaussian. UMAP normalizes away density differences. DensMAP restores some density information but severely distorts the Xs and scatters the I. ScaleMAP preserves all features. \emph{Bottom (Bridge):} two unequally sized Gaussians connected by a sparse linear bridge. UMAP compresses the bridge; DensMAP and ScaleMAP both recover it.}
\label{fig:synthetic}
\end{figure}

\subsection{Synthetic diagnostic}
\label{sec:synthetic}

Figure~\ref{fig:synthetic} illustrates the two failure modes that motivate ScaleMAP. On the XOI dataset---three X shapes of equal spatial extent but increasing point count, three Gaussians of equal count but increasing standard deviation, and one high-aspect-ratio Gaussian---UMAP renders the X shapes at sizes proportional to their point counts and the O shapes all at the same size, reflecting its density normalization. DensMAP restores the relative O sizes but enlarges and severely distorts the Xs; the high-aspect-ratio I sprays into nearby regions, with visible class mixing throughout. ScaleMAP correctly preserves equal X sizes, increasing O sizes, and the I shape, with no visible class mixing or disconnected points.
On the Bridge dataset---two unequally sized Gaussians connected by a sparse linear bridge---UMAP renders the clusters at the same size and compresses the bridge significantly. On clusters with more jagged edges, such as in transcriptomic data, such compressed bridges are easy to miss. A similar phenomenon occurs in hyperspectral images, where a sparse spectral spike can be compressed. DensMAP and ScaleMAP both produce correctly sized Gaussians and bridges.

\subsection{Standard benchmarks}

\begin{figure}[t]
\centering
\includegraphics[width=0.99\textwidth]{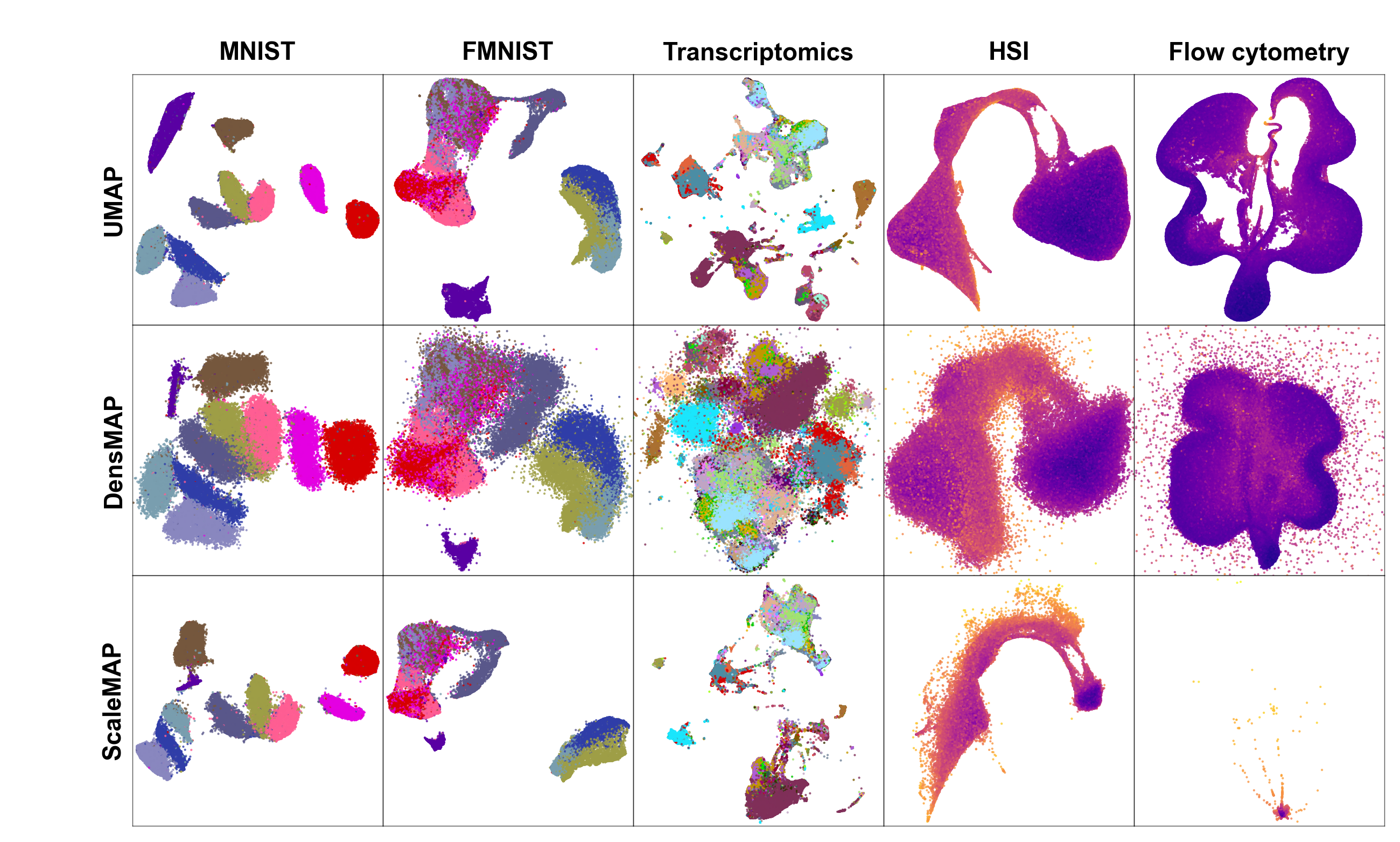}
\caption{\textbf{Embedding overview.} On MNIST and Fashion-MNIST, ScaleMAP closely resembles UMAP with clusters rescaled to reflect density. On the scientific datasets, DensMAP's tendency to scatter points is clearly visible, while ScaleMAP preserves density and reveals more underlying structure than UMAP. ScaleMAP's flow-cytometry embedding appears nearly empty at this scale, but this is not a failure: the multiscale structure becomes apparent upon zooming (Figure~\ref{fig:flow}).}
\label{fig:gallery}
\end{figure}

\begin{figure}[t]
\centering
\includegraphics[width=0.99\textwidth]{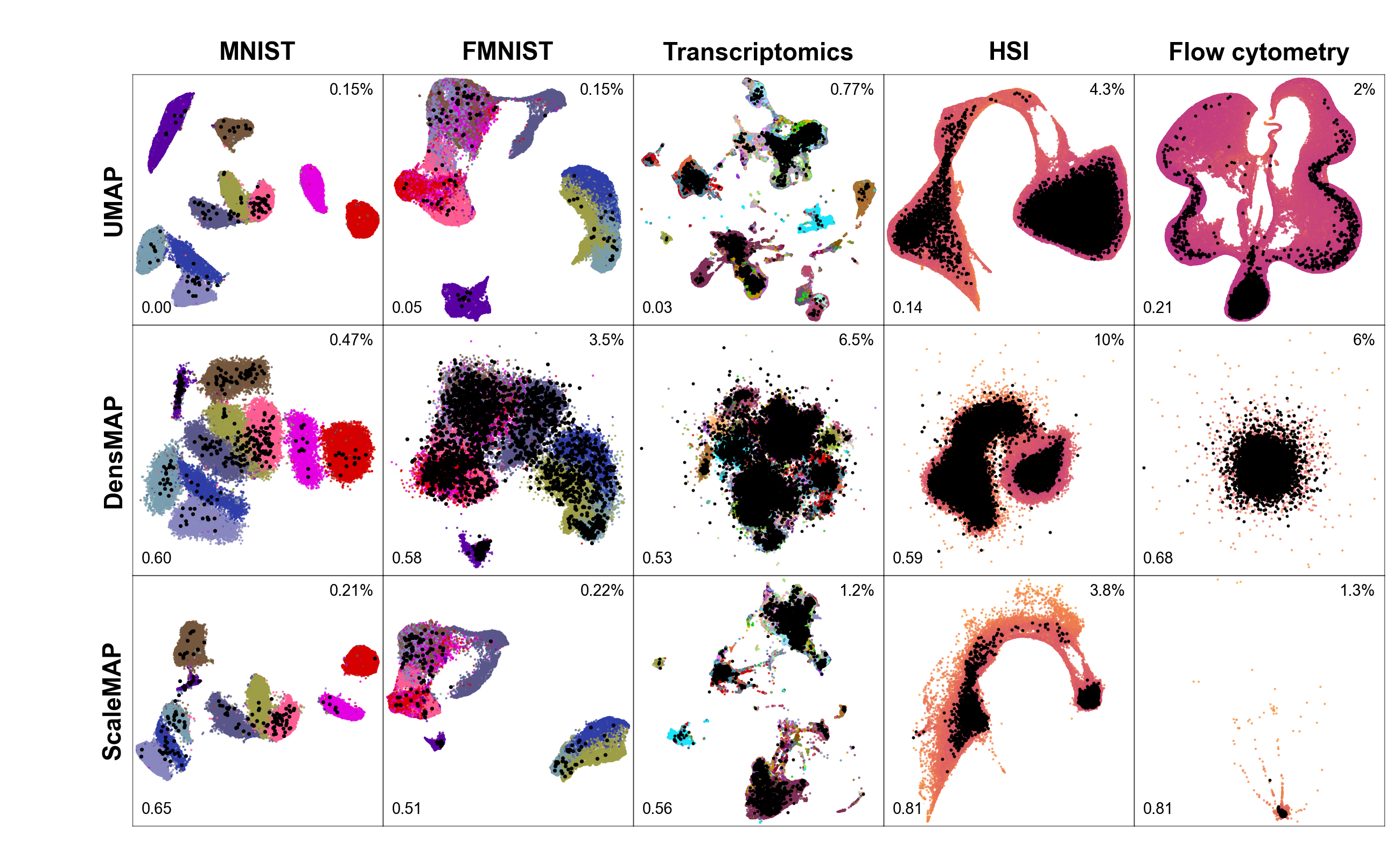}
\caption{\textbf{Disconnected points and density fits.} Same layout as Figure~\ref{fig:gallery}, with disconnected points overlaid in black. Inset numbers: bottom-left, density-preservation $R^2$; top-right, disconnected-point percentage. DensMAP produces 2--23x more disconnected points than UMAP. ScaleMAP's density $R^2$ is comparable to or exceeds DensMAP's on most datasets while keeping the disconnected fraction comparable to UMAP's.}
\label{fig:scatter-density}
\end{figure}

Across the four standard benchmarks (Figure~\ref{fig:gallery}, Figure~\ref{fig:scatter-density}, and Table~\ref{tab:benchmarks-no-error-bars}), we observe a consistent pattern. UMAP has very few disconnected points (e.g.\ 0.11\% on Fashion-MNIST) but low density preservation ($R^2 = 0.05$). DensMAP improves density preservation substantially ($R^2 = 0.58$) but at the cost of an order of magnitude more disconnected points (3.48\%), worse neighborhood preservation (recall 7.0\% vs UMAP's 13.3\%), and worse class mixing (47\% vs UMAP's 40\%). ScaleMAP attains density preservation comparable to DensMAP ($R^2 = 0.51$), neighborhood preservation comparable to UMAP (recall 14.3\%), disconnected-point fraction comparable to UMAP (0.19\%), and class mixing lower than either baseline (38\% vs UMAP 40\% and DensMAP 47\%). The pattern is qualitatively similar on MNIST, COIL-20, and Mammoth. On Mammoth, DensMAP obtains a significantly higher density $R^2$ than ScaleMAP (0.59 vs 0.39), but its embedding is visibly distorted (Figure~\ref{fig:all_dr_gallery}). ScaleMAP remains visually similar in quality to UMAP. 

\begin{figure}[t]
\centering
\includegraphics[width=1.0\textwidth]{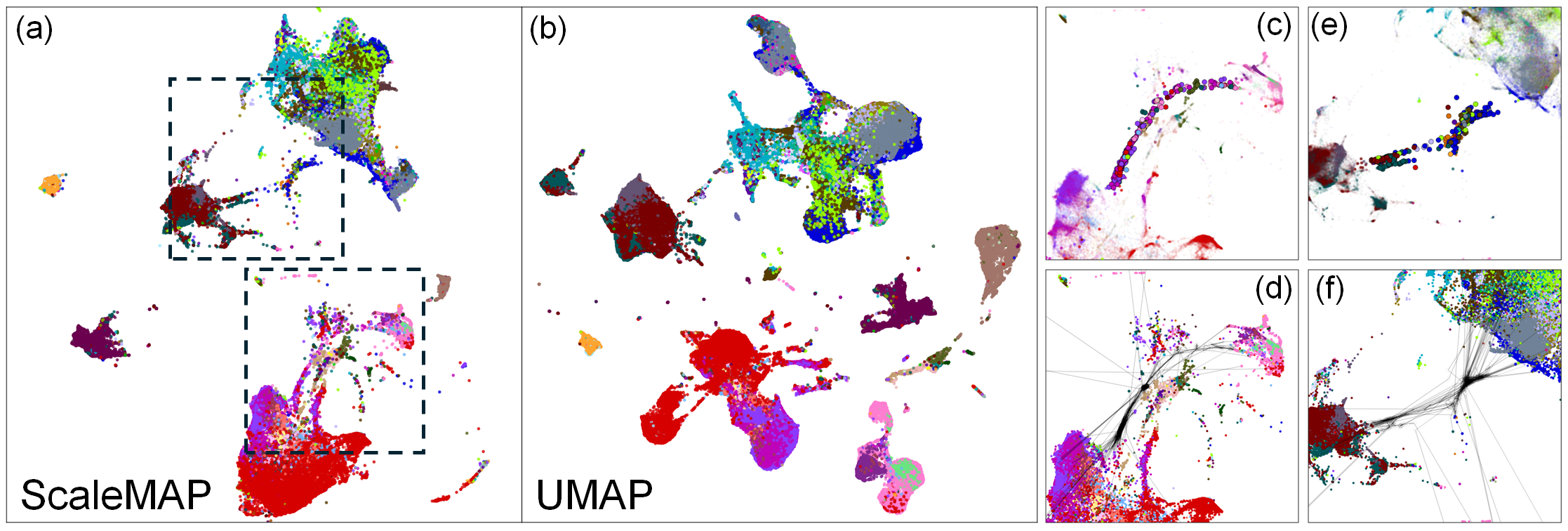}
\caption{\textbf{Transcriptomic bridges.} (a) ScaleMAP embedding of Tabula Sapiens immune cells; many bridges between cell types are visible. (b) UMAP embedding of the same data; these bridges are mostly absent. (c) Zoom on the bridge between monocytes (red/purple, bottom left) and neutrophils (pink, top right). (d) Network connectivity of (c), with edges drawn to each bridge point's two nearest original-space neighbors. (e) Zoom on the bridge between B cells (maroon/dark green, bottom left) and T cells (blue, top right). (f) Network connectivity of (e).}
\label{fig:transcriptomics}
\end{figure}

\subsection{Transcriptomics: recovery of sparse cell-state bridges}
\label{sec:transcriptomics}

In transcriptomic data, sparse bridges between well-defined cell populations correspond to cell-state transitions of biological interest. Figure~\ref{fig:transcriptomics} highlights two such bridges in the Tabula Sapiens immune-cell dataset that are recovered by ScaleMAP but invisible in the UMAP: the bridge between monocytes and neutrophils, which originate from myeloblasts, and the bridge between B and T cells, which originate from small lymphocytes. Both correspond to well-known developmental relationships, and we verify their connectivity by drawing edges from each bridge point to its two nearest neighbors in the original space (Figure~\ref{fig:transcriptomics}d, f). These bridges are visible in ScaleMAP across a range of neighborhood sizes $k$; in UMAP they are not visible across the full range we tested (k=10, 15, 30, 60). DensMAP's point scattering makes it difficult to determine whether potential bridges in its embedding correspond to real structure or to misplaced points. Several more bridges appear in the ScaleMAP. While verifying them is outside the scope of this work, these bridges could be used to identify candidate rare transitional cell states that are invisible to other embedding approaches.

\begin{figure}[t]
\centering
\includegraphics[width=1.0\textwidth]{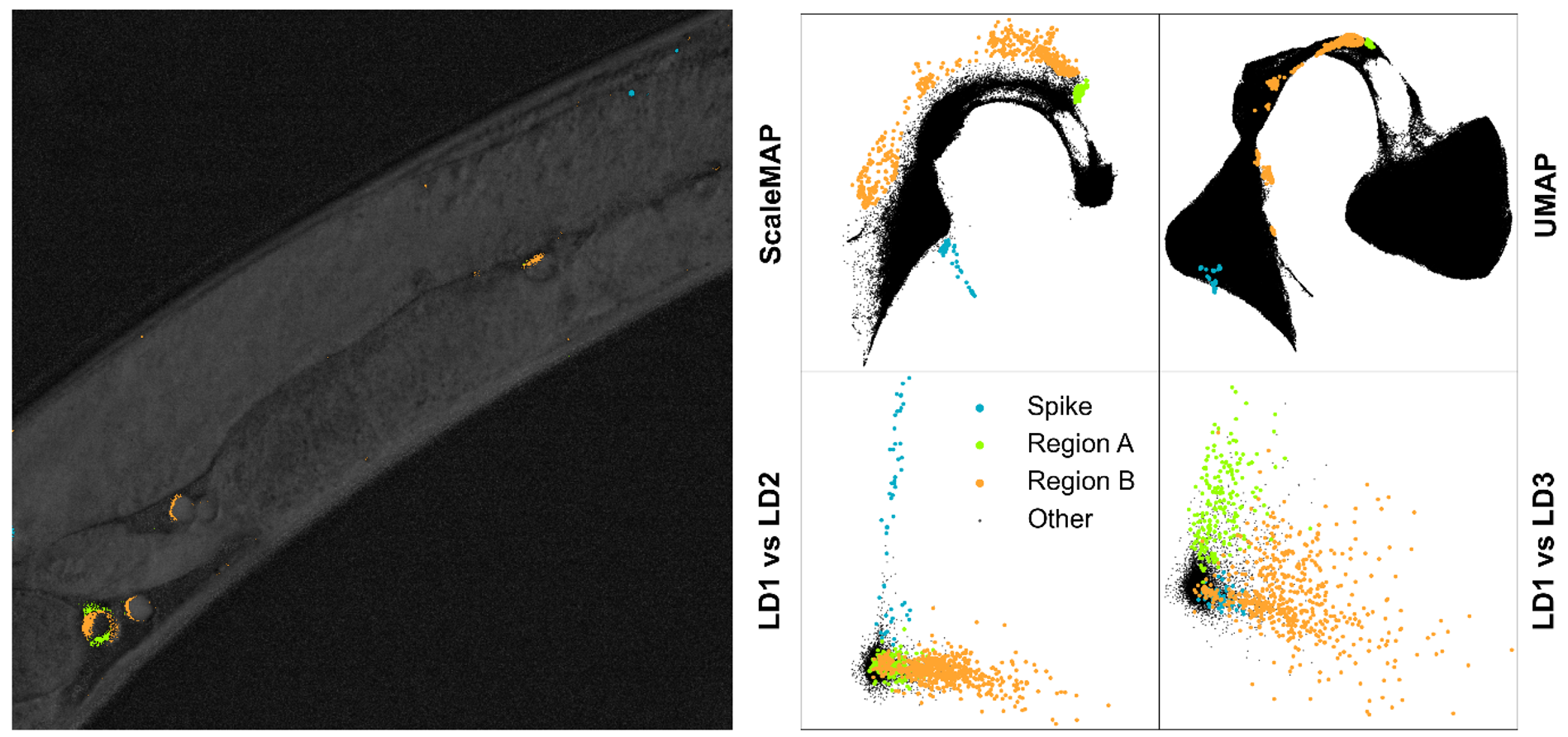}
\caption{\textbf{Hyperspectral imaging.} Left: spatial image of the \emph{C.~elegans} gonad with highlighted pixels corresponding to the annotated regions. Top: ScaleMAP and UMAP embeddings with three sparse structures annotated---a spectral spike, a small cluster (Region A), and a cloud of outlier pixels (Region B). UMAP compresses or absorbs all three. Bottom: linear discriminant projections confirm spectral separation of these regions from the bulk.}
\label{fig:hsi}
\end{figure}

\subsection{Hyperspectral imaging: recovery of sparse spectral spikes}
\label{sec:hsi}

Figure~\ref{fig:hsi} shows a coherent Raman image of a \emph{C.~elegans} gonad, embedded by ScaleMAP and UMAP. ScaleMAP reveals three kinds of structure that UMAP fails to surface: first, a sharp spike compressed to near-invisibility by UMAP---the direct analog of the bridge in Figure~\ref{fig:synthetic} (Spike); second, a small cluster (Region A) absorbed into the neighboring area in UMAP; and third, a diffuse cloud of outlier pixels (Region B) diverging from the main body. Linear discriminant analysis confirms that these regions are spectrally distinct. ScaleMAP also renders the large number of dark, background water pixels---the dominant population, occupying the right-center of both embeddings---at a size proportional to their (low) spectral variability, whereas UMAP inflates them to fill the embedding.

\begin{figure}[t]
\centering
\includegraphics[width=1.0\textwidth]{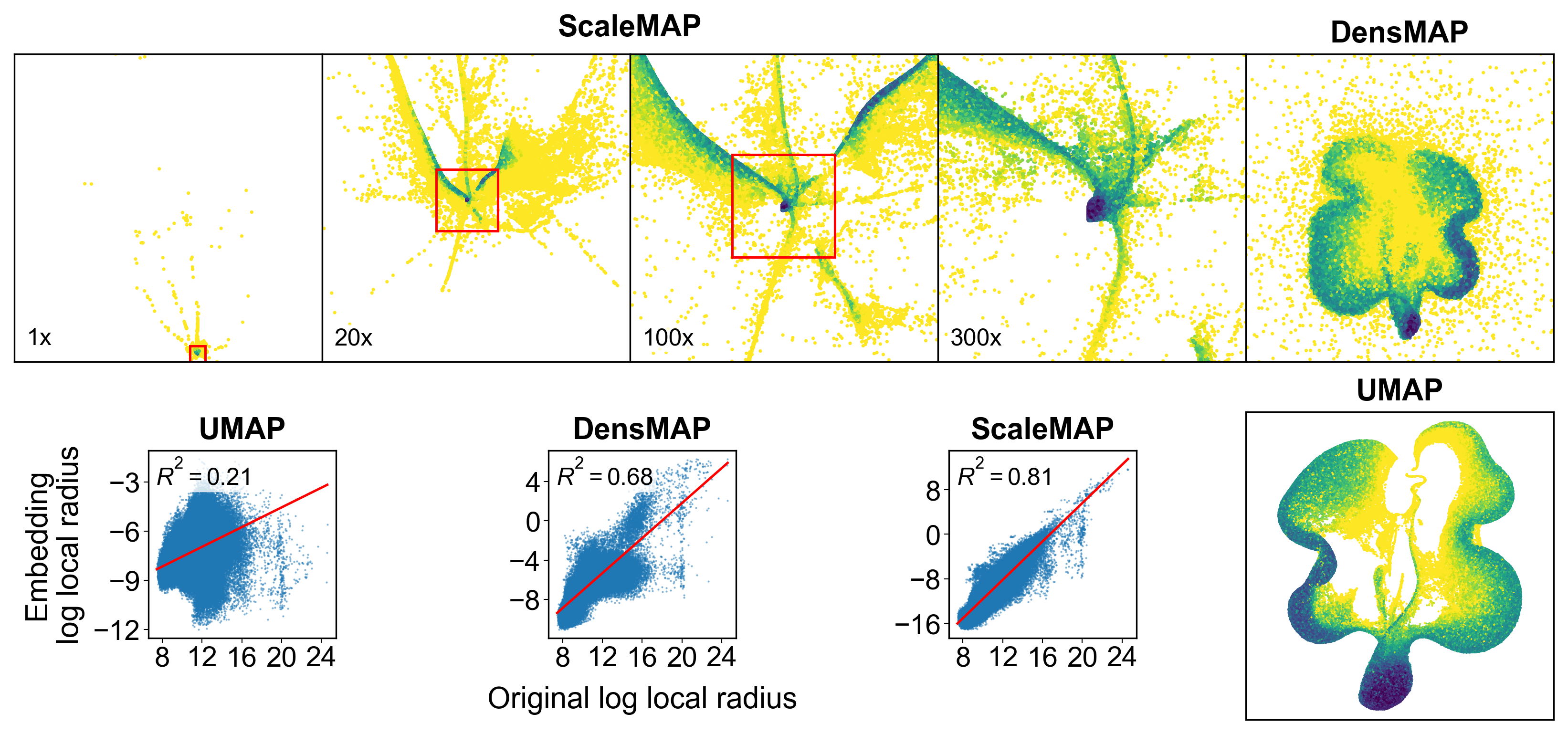}
\caption{\textbf{Flow cytometry: multiscale structure.} ScaleMAP embedding at four zoom levels, colored by log local radius, alongside DensMAP and UMAP embeddings at native scale. ScaleMAP preserves the multiscale structure of the data, and consequently requires $300\times$ zoom to resolve a dense, highly populated "bulb" of cells that are negative for all 8 markers. This bulb immediately stands out in both UMAP and DensMAP, reflecting their lack of multiscale structure. Bottom: log--log scatterplots of embedding versus original local radius. ScaleMAP maintains a tight linear relationship over 17 orders of magnitude ($R^2 = 0.81$), while DensMAP mixes different length-scales. See Figure~\ref{fig:flow_pca} for a PCA reflecting the multiscale structure.}
\label{fig:flow}
\end{figure}

\subsection{Flow cytometry: preserving multiscale structure}

Figure~\ref{fig:flow} shows ScaleMAP, DensMAP, and UMAP embeddings of bone-marrow flow-cytometry data, colored by log local radius. The ScaleMAP embedding is shown at four successive zoom levels ($1\times$, $20\times$, $100\times$, $300\times$). At $300\times$ magnification a dense central bulb becomes visible: a highly populated cluster of cells that are negative for all eight markers. This cluster is not rare---it accounts for a large fraction of the data---but its size in the embedding is tiny because the cells are indistinguishable under these markers. ScaleMAP preserves this multiscale structure faithfully: the log--log fit of local radius is approximately linear over 17 orders of magnitude ($R^2 = 0.81$). By contrast, UMAP and DensMAP look nearly identical on this dataset, and lack multiscale structure. DensMAP achieves an $R^2$ of 0.68, but the scatterplot reveals that it mixes different length scales---a consequence of the density term competing with the attractive forces rather than reshaping them.

\subsection{Portability to PaCMAP}

Applying the same change of variables to PaCMAP, Scale-PaCMAP improves density preservation relative to PaCMAP without dramatic degradation of neighborhood structure across all datasets tested (Figure~\ref{fig:all_dr_comparison}, Figure~\ref{fig:all_dr_gallery}, and Table~\ref{tab:supplement-metrics-density-preservation-r2}). The improvement is smaller than that from UMAP to ScaleMAP, but its consistency indicates that the underlying principle is not specific to UMAP. A fuller study, including Scale-PaCMAP's tendency to disconnect points that neither PaCMAP or ScaleMAP disconnects (Table~\ref{tab:supplement-metrics-disconnected-points}), particularly while "reducing" the 2-D synthetic benchmarks (such as with the XOI and Bridge datasets), is left to future work.

\begin{table}[t]
  \caption{\textbf{Benchmark results.} Values are means across five runs. U, D, and S denote UMAP, DensMAP, and ScaleMAP. Bold marks the best mean; underlining marks methods not significantly worse than the best under a paired one-sided $t$-test at $\alpha=0.05$. ScaleMAP achieves DensMAP-level density preservation (Dens. $R^2$), maintains UMAP-comparable neighborhood preservation (kNN Recall@15, disconnected-point percentage (Disconn. \%)), and also reduces class mixing (Mix \%). Dashes reflect the absence of class labels to calculate mixing. ScaleMAP runtime is roughly $2\times$ UMAP and comparable to DensMAP. Extended benchmarks with error bars in Section~\ref{sec:extended_benchmarks}}
  \label{tab:benchmarks-no-error-bars}
  \centering
  \small
  \setlength{\tabcolsep}{3pt}
  \begin{tabular}{lr@{.}lr@{.}lr@{.}lr@{.}lr@{.}lr@{.}lr@{.}lr@{.}lr@{.}lrrrr@{.}lr@{.}lr@{.}l}
    \toprule
    Dataset & \multicolumn{6}{c}{Recall@15} & \multicolumn{6}{c}{Disconn. \%} & \multicolumn{6}{c}{Dens. $R^2$} & \multicolumn{3}{c}{Mix \%} & \multicolumn{6}{c}{Runtime (s)} \\
    \cmidrule(lr){2-7} \cmidrule(lr){8-13} \cmidrule(lr){14-19} \cmidrule(lr){20-22} \cmidrule(lr){23-28}
     & \multicolumn{2}{c}{U} & \multicolumn{2}{c}{D} & \multicolumn{2}{c}{S} & \multicolumn{2}{c}{U} & \multicolumn{2}{c}{D} & \multicolumn{2}{c}{S} & \multicolumn{2}{c}{U} & \multicolumn{2}{c}{D} & \multicolumn{2}{c}{S} & \multicolumn{1}{c}{U} & \multicolumn{1}{c}{D} & \multicolumn{1}{c}{S} & \multicolumn{2}{c}{U} & \multicolumn{2}{c}{D} & \multicolumn{2}{c}{S} \\
    \midrule
    MNIST & 11 & 9 & 7 & 0 & \textbf{12} & \textbf{9} & \textbf{0} & \textbf{13} & 0 & 49 & 0 & 19 &  & 00 &  & 60 &  & \textbf{66} & 26 & 27 & \textbf{23} & \textbf{8} & \textbf{8} & 25 & 0 & 23 & 9 \\
    Fashion-MNIST & 13 & 3 & 7 & 1 & \textbf{14} & \textbf{3} & \textbf{0} & \textbf{11} & 3 & 48 & 0 & 19 &  & 05 &  & \textbf{58} &  & 51 & 40 & 47 & \textbf{38} & \textbf{11} & \textbf{2} & 29 & 0 & 27 & 7 \\
    COIL-20 & \textbf{76} & \textbf{5} & 66 & 5 & 74 & 1 & \underline{0} & \underline{00} & \underline{0} & \underline{00} & \textbf{0} & \textbf{00} &  & 09 &  & \textbf{66} &  & 61 & 7 & \underline{4} & \textbf{2} & \textbf{1} & \textbf{4} & 2 & 5 & 2 & 0 \\
    Mammoth & \textbf{70} & \textbf{7} & 51 & 0 & 68 & 8 & \underline{0} & \underline{00} & 0 & 03 & \textbf{0} & \textbf{00} &  & 00 &  & \textbf{59} &  & 39 & \textbf{1} & 13 & \underline{2} & \textbf{2} & \textbf{4} & 4 & 9 & 3 & 3 \\
    Transcriptomics & 11 & 7 & 4 & 6 & \textbf{12} & \textbf{4} & \textbf{0} & \textbf{81} & 6 & 37 & 1 & 12 &  & 03 &  & 54 &  & \textbf{55} & 37 & 48 & \textbf{34} & \textbf{69} & \textbf{9} & 136 & 3 & 131 & 7 \\
    HSI & 4 & 0 & 1 & 7 & \textbf{5} & \textbf{7} & 4 & 96 & 10 & 36 & \textbf{3} & \textbf{82} &  & 17 &  & 59 &  & \textbf{81} & \multicolumn{1}{c}{\textemdash{}} & \multicolumn{1}{c}{\textemdash{}} & \multicolumn{1}{c}{\textemdash{}} & \textbf{97} & \textbf{4} & 175 & 0 & 151 & 3 \\
    Flow cytometry & 11 & 4 & 3 & 1 & \textbf{15} & \textbf{0} & 2 & 08 & 6 & 13 & \textbf{1} & \textbf{35} &  & 20 &  & 68 &  & \textbf{80} & \multicolumn{1}{c}{\textemdash{}} & \multicolumn{1}{c}{\textemdash{}} & \multicolumn{1}{c}{\textemdash{}} & \textbf{421} & \textbf{7} & 567 & 4 & 598 & 2 \\
    \bottomrule
  \end{tabular}
\end{table}

\section{Ablations}
\label{sec:ablations}

We test two design choices in ScaleMAP: the symmetry of the rescaling, and its overall strength.

\begin{figure}[t]
\centering
\includegraphics[width=1.0\textwidth]{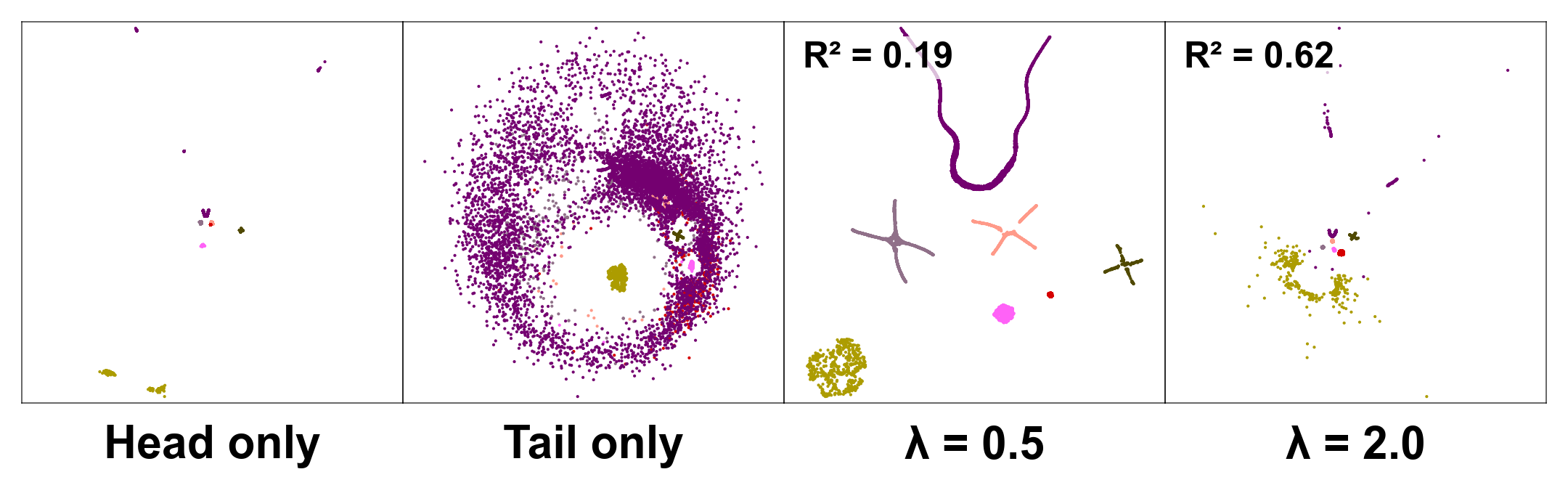}
\caption{\textbf{Ablations using XOI.} (a, b) Head-only and tail-only failure modes: replacing the geometric mean with one endpoint's radius fragments the embedding. (c) $\lambda = 0.5$: weakened rescaling, X-shape diagnostic fails. (d) $\lambda = 2$: over-rescaled, embedding breaks.}
\label{fig:ablations}
\end{figure}

\textbf{Both endpoints are required.} Replacing the geometric mean $\sqrt{r_i r_j}$ with either endpoint alone---that is, dividing the displacement only by $r_i$ or only by $r_j$ (the ``head-only'' and ``tail-only'' ablations)---causes the embedding to fragment. Neither asymmetric choice yields a usable embedding (Figure~\ref{fig:ablations}). The geometric mean is a natural symmetric choice that preserves the attraction--repulsion balance.

\textbf{Strength of the rescaling.} We expose a single coefficient $\lambda$ that interpolates between the original UMAP metric ($\lambda = 0$) and ScaleMAP ($\lambda = 1$); intermediate values weaken the rescaling and $\lambda > 1$ amplifies it. At $\lambda = 0.5$, the embedding looks qualitatively acceptable but the density $R^2$ is worse, and the X-shape diagnostic of Figure~\ref{fig:synthetic} fails (the X shapes regain their density-dependent sizing). At $\lambda = 2$, although the density $R^2$ is preserved, the embedding fragments.

\section{Discussion}
ScaleMAP performs well across datasets, both qualitatively and
quantitatively. It preserves neighborhood structure at least as well as
UMAP on five of seven datasets (recall@15;
Table~\ref{tab:benchmarks-no-error-bars}), with the two exceptions
within 3 percentage points and a comparable fraction of disconnected
points. DensMAP consistently has the weakest neighborhood preservation
of the three methods---roughly one third of UMAP's recall on the
scientific datasets, with 2--23$\times$ more disconnected points. Where
class labels are available, ScaleMAP produces the least artificial mixing
on every dataset. It also recovers sparse and multiscale structures not
visible in any of the other tested methods---the transcriptomic bridges
and spectral features of Sections~\ref{sec:transcriptomics}--\ref{sec:hsi}
are structures in existing, previously analyzed datasets.

At the same time, ScaleMAP's density preservation broadly matches
DensMAP's, and substantially exceeds it on the datasets with the widest
dynamic range of local radii ($R^2$ = 0.81 vs 0.59 on HSI; 0.80 vs 0.68
on flow cytometry). The one clear exception is Mammoth, where DensMAP
achieves a higher $R^2$ (0.59 vs 0.39) but at the cost of a visibly
distorted embedding (Figure~\ref{fig:all_dr_gallery}). 
The per-step cost of these improvements is negligible---O(1) arithmetic
per edge---though we default to $4\times$ UMAP's epochs, putting total
runtime at roughly $2\times$ UMAP and comparable to DensMAP.

We attribute this to how the two methods interact with UMAP's force
balance. DensMAP adds a density gradient only to the attractive update
(Eq.~\ref{eq:densmap-attr}--\ref{eq:densmap-rep}), so the density term
can oppose attraction while repulsion is unchanged---scattering points
away from their neighborhoods. ScaleMAP's rescaling enters both forces
symmetrically (Eq.~\ref{eq:scalemap-attr}--\ref{eq:scalemap-rep}),
shifting the equilibrium rather than breaking it. The change of variables approach is fundamentally more flexible than an additive penalty.

Surprisingly, ScaleMAP also improves global distance-order preservation
(Tables~\ref{tab:supplement-metrics-distance-spearman}--\ref{tab:supplement-metrics-random-triplet}),
despite not optimizing for it as PaCMAP does. The effect is largest
where the dynamic range of the local radii is widest: on flow cytometry,
ScaleMAP's pairwise distance-rank correlation (Spearman $\rho = 0.97$)
significantly exceeds both UMAP (0.74) and PaCMAP (0.78). 
This is consistent with a simple geometric intuition: structures
cannot be correctly arranged relative to each other when their
sizes are wrong.

\section{Limitations}
\label{sec:limitations}

ScaleMAP consistently produces more disconnected points than UMAP in some datasets (e.g., MNIST, FMNIST); the reason is not immediately clear, since, in principle, the change of variables preserves the attraction--repulsion balance. Increasing the number of optimization epochs reduces but does not eliminate the gap. On a synthetic ``beads-on-a-string'' dataset (Figure~\ref{fig:beads_on_string}), ScaleMAP compresses structure along the axis perpendicular to the string (Figure~\ref{fig:all_dr_gallery}), which we attribute to the local radius being a scalar representation of an anisotropic neighborhood. Global structure, although improved in ScaleMAP relative to UMAP, can be weaker than PaCMAP. See Appendix \ref{sec:extended_benchmarks}. On intrinsically two-dimensional synthetic datasets, ScaleMAP can require larger neighborhood sizes than UMAP to avoid fragmentation; this sensitivity was not observed in our higher-dimensional benchmarks (this issue is resolved even with the 3-D Mammoth dataset). Finally, many topologies cannot be faithfully represented in two dimensions, which is an issue shared by all the dimension reduction techniques discussed here. The structures ScaleMAP reveals are themselves more clearly visible in 3D. While our approach should in principle generalize to higher dimensions, we have not rigorously benchmarked this.

\section{Conclusion}
 
ScaleMAP re-injects local scale into UMAP through a change of variables
rather than a competing loss term. Across seven datasets spanning
standard benchmarks and three scientific modalities, it is the only
method tested that does not sacrifice one axis of embedding quality for
another: UMAP discards density, DensMAP degrades neighborhoods, and
PaCMAP does not address density at all. ScaleMAP combines their strengths---UMAP-level neighborhood preservation, DensMAP-level density
fidelity, and PaCMAP-competitive global order---and recovers structures that existing
methods miss, including transcriptomic bridges, sparse spectral
features, and density variation spanning 17 orders of magnitude. Restoring local scale improves global order: ScaleMAP narrows the gap to PaCMAP on most datasets, and exceeds it on two. Applied to PaCMAP, the same correction consistently restores density information, suggesting that local-scale correction is a general missing ingredient in neighborhood-preserving embeddings. These results make a case for ScaleMAP as the
default embedding method for visualizing high-dimensional data when faithfully representing scale matters.

\newpage

\section*{Acknowledgements}

This work was supported by the Department of Energy, Biological and Environmental Research (DOE BER DE-SC0022121). We thank Ronit
Sharon-Friling and Abigail Diering for their persistent use of early
versions of ScaleMAP in their research; their willingness
to adopt the method while it was still evolving motivated the
development of the version presented here.
 
\section*{Reproducibility statement}
 
All datasets are publicly available. Hyperparameters and preprocessing are described in Section~2, 3 and the appendix. We use five independent runs for all main results. Each run uses a distinct random seed, but multithreaded UMAP-family implementations are not bitwise deterministic, so we report mean and variation across runs rather than relying on exact reproducibility from a fixed seed. 

\section*{Broader impact}

We do not in general foresee specific negative applications for ScaleMAP beyond those associated with general-purpose data analysis. Because it is better than standard methods at surfacing small, rare subpopulations, it could in principle make it easier to visually isolate individuals in datasets where membership in a rare subgroup is itself identifying---a concern shared by any density-faithful visualization or analysis method. Identifying rare individuals or patterns may also be immediately beneficial to society if used to provide assistance or prevent fraud. 


\clearpage
\appendix

\section{Appendix}

\subsection{Reproducibility information}

$k=15$ nearest neighbors were used for all runs except for XOI, XO, and Square, for which $k=60$ nearest neighbors were used. Bridge used $k=90$ nearest neighbors, although equivalent results can be obtained with $k=60$ nearest neighbors. We set a random seed for each run to control initialization and stochastic sampling where supported. However, UMAP-family implementations are multithreaded and include nondeterministic parallel updates, so the same requested seed does not guarantee bitwise-identical embeddings. We therefore treat the five runs as independent repetitions under the standard multithreaded implementation and report means, with error bars in the appendix. All runs were on a Framework 16 laptop with 32 GB RAM with an AMD Ryzen AI 7 350 CPU. No thread count was specified during embedding - the embedding methods used all available CPU cores.

Although more powerful computers were used during iteration, this laptop would have been sufficient for executing the entire project, and would not have been much (>3x) slower than any other resources that were used. 

\subsection{Licenses and Terms of Use for Existing Assets}
\label{app:licenses}

This appendix summarizes the existing datasets and other assets used in this work, along with their licenses or applicable terms of use. We cite the original creators or maintainers of each asset in the main paper.

\begin{table}[h]
\caption{Summary of licenses and terms of use for existing assets used in this work.}
\centering
\small
\begin{tabular}{p{0.18\linewidth} p{0.27\linewidth} p{0.45\linewidth}}
\toprule
\textbf{Asset} & \textbf{License / Terms} & \textbf{Use in this work} \\
\midrule
MNIST & `Public' license. & We obtain the data from OpenML, which states that the license is `Public' and cite the GitHub source, which claims permission from the original source. \\

Fashion-MNIST & `Public' license. & We obtain the data from OpenML, which states that the license is `Public' and cite the original arXiv paper. \\

COIL-20 & Creative Commons Attribution 4.0 International & We obtain the data from OpenML and cite the original source. \\

Smithsonian Mammoth & Creative Commons Zero & We use the Smithsonian 3D \emph{Mammuthus primigenius} asset. The asset is made available under CC0 terms, which permit copying, modification, distribution, and use without restriction. We nevertheless cite the Smithsonian source for attribution and reproducibility. \\

Tabula Sapiens & We used processed data, which are licensed under CC BY 4.0. & We use the processed Tabula Sapiens data for academic analysis and cite the Tabula Sapiens Consortium. \\

SPADE cytometry data & Supplementary data are publicly downloadable from the Nature Biotechnology article page, but no separate open-data license is stated. & We use the supplementary flow-cytometry data associated with Qiu et al.'s SPADE paper for benchmarking and analysis. \\

Poorna et al. ACS JPCB 2023 & The Supporting Information is free to download, but no separate dataset license was identified. & We use the supplementary data associated with Poorna et al.'s spatial metabolomics study. We cite the original ACS paper. \\
\bottomrule
\end{tabular}
\label{tab:asset-licenses}
\end{table}

\subsection{Extended benchmarks}
\label{sec:extended_benchmarks}

Here, we introduce additional metrics for local neighborhood preservation and global distance-order preservation. The main text reports density preservation, disconnected points, artificial mixing, and runtime; the metrics below provide more ways of understanding the same neighborhood failures and also provide complementary measures of global structure.

Let $X=\{x_i\}_{i=1}^n$ denote the original data representation and
$Y=\{y_i\}_{i=1}^n$ the embedding. Let $\mathcal{N}_X^k(i)$ and
$\mathcal{N}_Y^k(i)$ denote the $k$ nearest neighbors of point $i$ in the
original and embedding spaces, respectively, excluding $i$ itself.

\paragraph{Neighborhood recall.}
We report neighborhood recall at $k=15$ and $k=100$:
\begin{equation}
\operatorname{Recall@}k =
\frac{1}{n}\sum_{i=1}^n
\frac{|\mathcal{N}_X^k(i)\cap \mathcal{N}_Y^k(i)|}{k}.
\label{eq:recall}
\end{equation}
This measures the fraction of original-space neighbors that remain neighbors in
the embedding.

\paragraph{Trustworthiness and continuity.}
Recall counts neighbor overlap but does not distinguish small rank errors from
catastrophic ones. We therefore also report trustworthiness and continuity at
$k=15$. Let $r_X(i,j)$ be the rank of $j$ by distance from $x_i$ in the
original space, with rank $1$ denoting the nearest non-self neighbor, and define
$U_i^k=\mathcal{N}_Y^k(i)\setminus \mathcal{N}_X^k(i)$. The standard
trustworthiness score is
\begin{equation}
\operatorname{Trust}(k)
=
1-
\frac{2}{n k(2n-3k-1)}
\sum_{i=1}^n
\sum_{j\in U_i^k}
(r_X(i,j)-k).
\label{eq:trustworthiness}
\end{equation}
It penalizes points that appear close in the embedding despite being far away in
the original space. For computational efficiency on large datasets, we estimate
Eq.~\ref{eq:trustworthiness} on a fixed set $\mathcal{A}$ of
$\min(10{,}000,n)$ anchor points sampled once per dataset and reused across all embedding methods and runs:
\begin{equation}
\widehat{\operatorname{Trust}}(k)
=
1-
\frac{2}{|\mathcal{A}| k(2n-3k-1)}
\sum_{i\in \mathcal{A}}
\sum_{j\in U_i^k}
(r_X(i,j)-k).
\label{eq:trustworthiness-sampled}
\end{equation}
For these sampled anchors, ranks $r_X(i,j)$ are computed exactly by distance
counting in the original space; only the anchor set is sampled.

Continuity measures the complementary failure mode. Let $r_Y(i,j)$ be the rank of $j$ by distance
from $y_i$ in the embedding, and define
$V_i^k=\mathcal{N}_X^k(i)\setminus \mathcal{N}_Y^k(i)$. Continuity is
\begin{equation}
\operatorname{Cont}(k)
=
1-
\frac{2}{n k(2n-3k-1)}
\sum_{i=1}^n
\sum_{j\in V_i^k}
(r_Y(i,j)-k).
\label{eq:continuity}
\end{equation}
It penalizes original-space neighbors that are separated in the embedding. We
compute continuity exactly over all points using embedding-space ranks. Both
trustworthiness and continuity lie in $[0,1]$, with larger values indicating
better neighborhood preservation.

\paragraph{Global distance-order metrics.}
We report two rank-based proxies for global structure. Random triplet accuracy
samples triples $(i,j,\ell)$ and measures whether the embedding preserves the
relative ordering of distances from the anchor:
\begin{equation}
\operatorname{Triplet}
=
\frac{1}{M}
\sum_{m=1}^{M}
\mathbf{1}
\!\left[
\left(d_X(i_m,j_m)<d_X(i_m,\ell_m)\right)
=
\left(d_Y(i_m,j_m)<d_Y(i_m,\ell_m)\right)
\right].
\label{eq:triplet}
\end{equation}
Distance Spearman samples $\min(10^6,{n \choose 2})$ unordered pairs $(i,j)$,
forms vectors $a_m=d_X(i_m,j_m)$ and $b_m=d_Y(i_m,j_m)$, and reports the
Spearman rank correlation $\rho_s(a,b)$. The same sampled pairs are reused
across methods and runs for a given dataset. Triplet accuracy measures
anchor-relative distance ordering, whereas Distance Spearman measures pairwise
distance-rank preservation over sampled pairs. Larger values indicate better
global distance-order preservation.

Table values are mean $\pm$ two standard deviations across five runs. Bold marks the best mean; underlining marks methods not significantly worse than the best under a paired one-sided $t$-test at $\alpha=0.05$.

Here, we compare ScaleMAP at both its default 800 epochs as well as when reduced to UMAP's default 200 epochs. We also compare against UMAP and DensMAP when given 800 epochs. DensMAP is also used at its default epochs. This way, we get to compare these three methods at approximately equal compute and runtime. PaCMAP is used at its default (100, 100, 250) epochs, while Scale-PaCMAP is given twice the epochs of PaCMAP, at (200, 200, 500). 

Broadly, we note that the even when UMAP and DensMAP are matched to 800 epochs, the pattern reported in the main paper holds. ScaleMAP performs surprisingly well in the flow cytometry dataset, even exceeding PaCMAP's global structure metrics despite not actively optimizing for global structure, highlighting the importance of conserving scale in multiscale datasets. We suspect that Scale-PaCMAP underperforms at multiple metrics because the mid-near points use the same change of variables as the near and far points, which is likely inappropriate. Nonetheless, it outperforms all other tested methods on at least some datasets on multiple benchmarks, and generally shows performance around the median for most metrics.

\begin{table}[h]
  \caption{Disconnected points (\%). }
  \label{tab:supplement-metrics-disconnected-points}
  \centering
  \small
  \setlength{\tabcolsep}{3pt}
  \makebox[\textwidth][c]{%
  \begin{tabular}{lccccccc}
    \toprule
    Embedding & MNIST & FMNIST & COIL-20 & Mammoth & Transcriptomics & HSI & Flow cytometry \\
    \midrule
    ScaleMAP & 0.19 $\pm$ 0.05 & 0.19 $\pm$ 0.03 & \textbf{0.00 $\pm$ 0.00} & \textbf{0.00 $\pm$ 0.00} & 1.12 $\pm$ 0.05 & \textbf{3.82 $\pm$ 0.04} & \textbf{1.35 $\pm$ 0.01} \\
    ScaleMAP-200 & 0.22 $\pm$ 0.03 & 0.23 $\pm$ 0.04 & \underline{0.00 $\pm$ 0.00} & \underline{0.00 $\pm$ 0.00} & 1.19 $\pm$ 0.03 & 4.54 $\pm$ 0.06 & 1.80 $\pm$ 0.03 \\
    UMAP & 0.13 $\pm$ 0.03 & 0.11 $\pm$ 0.01 & \underline{0.00 $\pm$ 0.00} & \underline{0.00 $\pm$ 0.00} & 0.81 $\pm$ 0.05 & 4.96 $\pm$ 1.71 & 2.08 $\pm$ 0.04 \\
    UMAP-800 & \textbf{0.12 $\pm$ 0.02} & \textbf{0.10 $\pm$ 0.01} & \underline{0.00 $\pm$ 0.00} & \underline{0.00 $\pm$ 0.00} & \textbf{0.74 $\pm$ 0.05} & 4.00 $\pm$ 0.12 & 2.01 $\pm$ 0.02 \\
    DensMAP & 0.49 $\pm$ 0.06 & 3.48 $\pm$ 0.11 & \underline{0.00 $\pm$ 0.00} & 0.03 $\pm$ 0.04 & 6.37 $\pm$ 0.05 & 10.36 $\pm$ 0.13 & 6.13 $\pm$ 0.08 \\
    DensMAP-800 & 0.29 $\pm$ 0.06 & 2.60 $\pm$ 0.05 & \underline{0.00 $\pm$ 0.00} & 0.03 $\pm$ 0.03 & 4.35 $\pm$ 0.22 & 8.96 $\pm$ 0.05 & 4.61 $\pm$ 0.04 \\
    PaCMAP & 0.34 $\pm$ 0.03 & 0.36 $\pm$ 0.03 & \underline{0.00 $\pm$ 0.00} & \underline{0.00 $\pm$ 0.00} & 1.55 $\pm$ 0.08 & 5.40 $\pm$ 0.09 & 2.50 $\pm$ 0.04 \\
    Scale-PaCMAP & 0.46 $\pm$ 0.06 & 0.45 $\pm$ 0.05 & \underline{0.00 $\pm$ 0.00} & \underline{0.00 $\pm$ 0.00} & 1.76 $\pm$ 0.07 & 5.80 $\pm$ 0.18 & 2.52 $\pm$ 0.04 \\
    \bottomrule
  \end{tabular}
  }
\end{table}

\begin{table}[h]
  \caption{Density preservation ($R^2$).}
  \label{tab:supplement-metrics-density-preservation-r2}
  \centering
  \small
  \setlength{\tabcolsep}{3pt}
  \makebox[\textwidth][c]{%
  \begin{tabular}{lccccccc}
    \toprule
    Embedding & MNIST & FMNIST & COIL-20 & Mammoth & Transcriptomics & HSI & Flow cytometry \\
    \midrule
    ScaleMAP & \textbf{.661 $\pm$ .010} & .510 $\pm$ .001 & .612 $\pm$ .059 & .386 $\pm$ .014 & .547 $\pm$ .014 & \textbf{.805 $\pm$ .001} & .797 $\pm$ .003 \\
    ScaleMAP-200 & .629 $\pm$ .011 & .498 $\pm$ .004 & .595 $\pm$ .040 & .354 $\pm$ .012 & .522 $\pm$ .012 & .797 $\pm$ .003 & \textbf{.815 $\pm$ .002} \\
    UMAP & .004 $\pm$ .003 & .052 $\pm$ .009 & .094 $\pm$ .036 & .003 $\pm$ .002 & .034 $\pm$ .014 & .167 $\pm$ .053 & .203 $\pm$ .007 \\
    UMAP-800 & .003 $\pm$ .002 & .051 $\pm$ .007 & .085 $\pm$ .027 & .004 $\pm$ .004 & .029 $\pm$ .005 & .115 $\pm$ .012 & .178 $\pm$ .009 \\
    DensMAP & .605 $\pm$ .025 & .579 $\pm$ .011 & \underline{.658 $\pm$ .049} & \textbf{.587 $\pm$ .017} & .537 $\pm$ .007 & .590 $\pm$ .003 & .677 $\pm$ .004 \\
    DensMAP-800 & .630 $\pm$ .007 & \textbf{.605 $\pm$ .005} & \textbf{.665 $\pm$ .019} & \underline{.577 $\pm$ .049} & \textbf{.576 $\pm$ .010} & .634 $\pm$ .001 & .698 $\pm$ .006 \\
    PaCMAP & .002 $\pm$ .001 & .063 $\pm$ .007 & .145 $\pm$ .088 & .001 $\pm$ .001 & .037 $\pm$ .006 & .146 $\pm$ .014 & .180 $\pm$ .033 \\
    Scale-PaCMAP & .437 $\pm$ .009 & .311 $\pm$ .008 & .641 $\pm$ .027 & .103 $\pm$ .042 & .282 $\pm$ .014 & .672 $\pm$ .028 & .746 $\pm$ .004 \\
    \bottomrule
  \end{tabular}
  }
\end{table}

\begin{table}[h]
  \caption{Artificial mixing (\%)}
  \label{tab:supplement-metrics-artificial-mixing}
  \centering
  \small
  \setlength{\tabcolsep}{3pt}
  \makebox[\textwidth][c]{%
  \begin{tabular}{lccccccc}
    \toprule
    Embedding & MNIST & FMNIST & COIL-20 & Mammoth & Transcriptomics & HSI & Flow cytometry \\
    \midrule
    ScaleMAP & 22.62 $\pm$ 0.33 & \textbf{38.22 $\pm$ 0.99} & \textbf{1.84 $\pm$ 1.41} & \underline{1.74 $\pm$ 0.64} & \textbf{34.03 $\pm$ 0.71} & \textemdash{} & \textemdash{} \\
    ScaleMAP-200 & 23.18 $\pm$ 0.55 & \underline{38.74 $\pm$ 0.81} & 6.73 $\pm$ 2.16 & 1.74 $\pm$ 0.29 & 35.13 $\pm$ 0.52 & \textemdash{} & \textemdash{} \\
    UMAP & 26.44 $\pm$ 0.23 & 40.47 $\pm$ 0.85 & 6.73 $\pm$ 3.00 & \textbf{1.45 $\pm$ 0.30} & 36.62 $\pm$ 0.26 & \textemdash{} & \textemdash{} \\
    UMAP-800 & 26.16 $\pm$ 0.62 & 40.28 $\pm$ 1.43 & 6.40 $\pm$ 2.33 & 1.75 $\pm$ 0.37 & 36.26 $\pm$ 0.51 & \textemdash{} & \textemdash{} \\
    DensMAP & 27.47 $\pm$ 2.14 & 46.80 $\pm$ 0.93 & \underline{3.73 $\pm$ 3.99} & 13.41 $\pm$ 2.64 & 47.98 $\pm$ 0.87 & \textemdash{} & \textemdash{} \\
    DensMAP-800 & 26.78 $\pm$ 0.95 & 44.53 $\pm$ 0.40 & \underline{4.90 $\pm$ 7.35} & 11.78 $\pm$ 3.33 & 45.19 $\pm$ 1.32 & \textemdash{} & \textemdash{} \\
    PaCMAP & 20.44 $\pm$ 0.66 & 40.71 $\pm$ 0.81 & 6.54 $\pm$ 2.35 & 5.94 $\pm$ 1.99 & 37.22 $\pm$ 0.18 & \textemdash{} & \textemdash{} \\
    Scale-PaCMAP & \textbf{19.57 $\pm$ 0.27} & 38.69 $\pm$ 0.72 & 4.70 $\pm$ 3.38 & 3.73 $\pm$ 4.02 & 37.64 $\pm$ 0.59 & \textemdash{} & \textemdash{} \\
    \bottomrule
  \end{tabular}
  }
\end{table}

\begin{table}[h]
  \caption{KNN recall@15 (\%)}
  \label{tab:supplement-metrics-knn-recall-15}
  \centering
  \small
  \setlength{\tabcolsep}{3pt}
  \makebox[\textwidth][c]{%
  \begin{tabular}{lccccccc}
    \toprule
    Embedding & MNIST & FMNIST & COIL-20 & Mammoth & Transcriptomics & HSI & Flow cytometry \\
    \midrule
    ScaleMAP & \textbf{12.89 $\pm$ 0.15} & \textbf{14.26 $\pm$ 0.01} & 74.11 $\pm$ 0.65 & 68.80 $\pm$ 0.57 & 12.41 $\pm$ 0.18 & \textbf{5.66 $\pm$ 0.03} & \textbf{15.02 $\pm$ 0.07} \\
    ScaleMAP-200 & 11.76 $\pm$ 0.19 & 13.08 $\pm$ 0.13 & 71.74 $\pm$ 1.14 & 66.91 $\pm$ 0.57 & 11.17 $\pm$ 0.24 & 4.81 $\pm$ 0.06 & 12.14 $\pm$ 0.08 \\
    UMAP & 11.89 $\pm$ 0.06 & 13.27 $\pm$ 0.25 & \underline{76.49 $\pm$ 0.89} & 70.71 $\pm$ 0.39 & 11.71 $\pm$ 0.16 & 3.99 $\pm$ 0.26 & 11.41 $\pm$ 0.08 \\
    UMAP-800 & 12.62 $\pm$ 0.08 & 14.08 $\pm$ 0.13 & \textbf{76.88 $\pm$ 0.63} & \textbf{71.11 $\pm$ 0.49} & \textbf{12.54 $\pm$ 0.11} & 4.27 $\pm$ 0.03 & 11.79 $\pm$ 0.10 \\
    DensMAP & 7.04 $\pm$ 0.07 & 7.09 $\pm$ 0.06 & 66.53 $\pm$ 1.81 & 50.99 $\pm$ 0.97 & 4.59 $\pm$ 0.07 & 1.70 $\pm$ 0.02 & 3.15 $\pm$ 0.04 \\
    DensMAP-800 & 8.23 $\pm$ 0.13 & 7.92 $\pm$ 0.04 & 67.67 $\pm$ 1.54 & 52.13 $\pm$ 0.54 & 5.75 $\pm$ 0.09 & 1.93 $\pm$ 0.03 & 3.98 $\pm$ 0.07 \\
    PaCMAP & 8.97 $\pm$ 0.08 & 9.27 $\pm$ 0.05 & 70.04 $\pm$ 0.69 & 60.96 $\pm$ 0.92 & 8.16 $\pm$ 0.08 & 3.01 $\pm$ 0.04 & 8.90 $\pm$ 0.10 \\
    Scale-PaCMAP & 9.16 $\pm$ 0.23 & 9.90 $\pm$ 0.17 & 71.23 $\pm$ 0.90 & 64.55 $\pm$ 2.49 & 8.82 $\pm$ 0.04 & 3.06 $\pm$ 0.07 & 9.02 $\pm$ 0.05 \\
    \bottomrule
  \end{tabular}
  }
\end{table}

\begin{table}[h]
  \caption{KNN recall@100 (\%)}
  \label{tab:supplement-metrics-knn-recall-100}
  \centering
  \small
  \setlength{\tabcolsep}{3pt}
  \makebox[\textwidth][c]{%
  \begin{tabular}{lccccccc}
    \toprule
    Embedding & MNIST & FMNIST & COIL-20 & Mammoth & Transcriptomics & HSI & Flow cytometry \\
    \midrule
    ScaleMAP & 23.05 $\pm$ 0.25 & \textbf{25.81 $\pm$ 0.03} & \textbf{57.35 $\pm$ 0.95} & 71.87 $\pm$ 0.31 & 22.62 $\pm$ 0.30 & \textbf{9.39 $\pm$ 0.03} & 19.71 $\pm$ 0.07 \\
    ScaleMAP-200 & 22.34 $\pm$ 0.32 & 24.98 $\pm$ 0.18 & \underline{56.82 $\pm$ 1.56} & 71.05 $\pm$ 0.53 & 21.75 $\pm$ 0.34 & 8.77 $\pm$ 0.08 & 18.75 $\pm$ 0.09 \\
    UMAP & 22.88 $\pm$ 0.06 & 25.14 $\pm$ 0.50 & 56.08 $\pm$ 1.66 & \underline{73.19 $\pm$ 0.40} & 22.54 $\pm$ 0.23 & 8.17 $\pm$ 0.50 & 20.24 $\pm$ 0.21 \\
    UMAP-800 & \textbf{23.43 $\pm$ 0.10} & 25.72 $\pm$ 0.11 & 55.89 $\pm$ 0.59 & \textbf{73.68 $\pm$ 1.26} & \textbf{23.10 $\pm$ 0.11} & 8.53 $\pm$ 0.04 & \textbf{20.47 $\pm$ 0.23} \\
    DensMAP & 14.69 $\pm$ 0.20 & 16.43 $\pm$ 0.09 & 55.51 $\pm$ 1.25 & 60.51 $\pm$ 1.70 & 11.11 $\pm$ 0.07 & 4.15 $\pm$ 0.01 & 7.85 $\pm$ 0.09 \\
    DensMAP-800 & 16.21 $\pm$ 0.31 & 17.56 $\pm$ 0.04 & 55.62 $\pm$ 2.17 & 61.57 $\pm$ 0.84 & 12.86 $\pm$ 0.17 & 4.51 $\pm$ 0.06 & 9.18 $\pm$ 0.19 \\
    PaCMAP & 21.28 $\pm$ 0.12 & 21.55 $\pm$ 0.12 & \underline{57.08 $\pm$ 1.39} & 71.03 $\pm$ 0.74 & 18.24 $\pm$ 0.13 & 7.48 $\pm$ 0.06 & 18.38 $\pm$ 0.19 \\
    Scale-PaCMAP & 21.48 $\pm$ 0.56 & 22.47 $\pm$ 0.33 & \underline{57.32 $\pm$ 0.83} & \underline{72.94 $\pm$ 3.58} & 19.16 $\pm$ 0.07 & 7.48 $\pm$ 0.11 & 18.58 $\pm$ 0.07 \\
    \bottomrule
  \end{tabular}
  }
\end{table}

\begin{table}[h]
  \caption{KNN continuity@15}
  \label{tab:supplement-metrics-knn-continuity-15}
  \centering
  \small
  \setlength{\tabcolsep}{3pt}
  \makebox[\textwidth][c]{%
  \begin{tabular}{lccccccc}
    \toprule
    Embedding & MNIST & FMNIST & COIL-20 & Mammoth & Transcriptomics & HSI & Flow cytometry \\
    \midrule
    ScaleMAP & \textbf{.984 $\pm$ .000} & .991 $\pm$ .000 & .993 $\pm$ .001 & .997 $\pm$ .000 & .996 $\pm$ .000 & \textbf{.995 $\pm$ .000} & .998 $\pm$ .000 \\
    ScaleMAP-200 & .983 $\pm$ .000 & .991 $\pm$ .000 & .992 $\pm$ .002 & .998 $\pm$ .000 & .995 $\pm$ .000 & .995 $\pm$ .000 & .998 $\pm$ .000 \\
    UMAP & .983 $\pm$ .000 & \textbf{.991 $\pm$ .000} & \textbf{.994 $\pm$ .000} & .998 $\pm$ .001 & .995 $\pm$ .000 & .994 $\pm$ .002 & .998 $\pm$ .000 \\
    UMAP-800 & .983 $\pm$ .000 & \underline{.991 $\pm$ .000} & .993 $\pm$ .001 & .998 $\pm$ .001 & .995 $\pm$ .000 & .995 $\pm$ .000 & .998 $\pm$ .000 \\
    DensMAP & .982 $\pm$ .000 & .988 $\pm$ .000 & .992 $\pm$ .001 & .996 $\pm$ .000 & .993 $\pm$ .000 & .993 $\pm$ .000 & .993 $\pm$ .000 \\
    DensMAP-800 & .983 $\pm$ .000 & .989 $\pm$ .000 & .992 $\pm$ .001 & .996 $\pm$ .000 & .994 $\pm$ .001 & .994 $\pm$ .000 & .995 $\pm$ .000 \\
    PaCMAP & .980 $\pm$ .001 & .990 $\pm$ .000 & .989 $\pm$ .001 & \textbf{.999 $\pm$ .000} & .996 $\pm$ .000 & .994 $\pm$ .000 & .998 $\pm$ .000 \\
    Scale-PaCMAP & .981 $\pm$ .000 & .990 $\pm$ .000 & .991 $\pm$ .001 & \underline{.998 $\pm$ .001} & \textbf{.996 $\pm$ .000} & .995 $\pm$ .001 & \textbf{.998 $\pm$ .000} \\
    \bottomrule
  \end{tabular}
  }
\end{table}

\begin{table}[h]
  \caption{KNN trustworthiness@15}
  \label{tab:supplement-metrics-knn-trustworthiness-15}
  \centering
  \small
  \setlength{\tabcolsep}{3pt}
  \makebox[\textwidth][c]{%
  \begin{tabular}{lccccccc}
    \toprule
    Embedding & MNIST & FMNIST & COIL-20 & Mammoth & Transcriptomics & HSI & Flow cytometry \\
    \midrule
    ScaleMAP & \underline{.960 $\pm$ .002} & \textbf{.979 $\pm$ .000} & \textbf{.994 $\pm$ .001} & .999 $\pm$ .000 & \underline{.987 $\pm$ .001} & .964 $\pm$ .000 & .988 $\pm$ .000 \\
    ScaleMAP-200 & .958 $\pm$ .002 & .978 $\pm$ .000 & .992 $\pm$ .001 & .999 $\pm$ .000 & .987 $\pm$ .000 & .963 $\pm$ .000 & .986 $\pm$ .001 \\
    UMAP & .958 $\pm$ .000 & .977 $\pm$ .003 & .992 $\pm$ .002 & \underline{.999 $\pm$ .000} & \underline{.988 $\pm$ .001} & .959 $\pm$ .013 & .988 $\pm$ .000 \\
    UMAP-800 & \textbf{.960 $\pm$ .001} & .978 $\pm$ .000 & .992 $\pm$ .001 & \textbf{.999 $\pm$ .000} & \textbf{.988 $\pm$ .001} & \textbf{.965 $\pm$ .000} & \textbf{.988 $\pm$ .000} \\
    DensMAP & .928 $\pm$ .002 & .951 $\pm$ .001 & .969 $\pm$ .006 & .991 $\pm$ .003 & .963 $\pm$ .001 & .956 $\pm$ .000 & .960 $\pm$ .001 \\
    DensMAP-800 & .935 $\pm$ .004 & .956 $\pm$ .001 & .971 $\pm$ .011 & .993 $\pm$ .001 & .968 $\pm$ .003 & .957 $\pm$ .000 & .968 $\pm$ .001 \\
    PaCMAP & .953 $\pm$ .002 & .971 $\pm$ .001 & .986 $\pm$ .004 & .993 $\pm$ .002 & .980 $\pm$ .001 & .963 $\pm$ .000 & .986 $\pm$ .001 \\
    Scale-PaCMAP & .951 $\pm$ .002 & .973 $\pm$ .001 & .985 $\pm$ .005 & .996 $\pm$ .005 & .980 $\pm$ .001 & .963 $\pm$ .001 & .985 $\pm$ .000 \\
    \bottomrule
  \end{tabular}
  }
\end{table}

\begin{table}[h]
  \caption{Distance Spearman}
  \label{tab:supplement-metrics-distance-spearman}
  \centering
  \small
  \setlength{\tabcolsep}{3pt}
  \makebox[\textwidth][c]{%
  \begin{tabular}{lccccccc}
    \toprule
    Embedding & MNIST & FMNIST & COIL-20 & Mammoth & Transcriptomics & HSI & Flow cytometry \\
    \midrule
    ScaleMAP & .397 $\pm$ .033 & .588 $\pm$ .005 & .271 $\pm$ .053 & .836 $\pm$ .007 & .579 $\pm$ .025 & .931 $\pm$ .010 & \textbf{.967 $\pm$ .004} \\
    ScaleMAP-200 & .392 $\pm$ .020 & \textbf{.615 $\pm$ .010} & .275 $\pm$ .085 & .809 $\pm$ .018 & .549 $\pm$ .028 & .868 $\pm$ .005 & .938 $\pm$ .006 \\
    UMAP & .302 $\pm$ .013 & .581 $\pm$ .011 & .245 $\pm$ .059 & .815 $\pm$ .023 & .535 $\pm$ .026 & .820 $\pm$ .009 & .738 $\pm$ .009 \\
    UMAP-800 & .318 $\pm$ .025 & .559 $\pm$ .005 & .227 $\pm$ .123 & .823 $\pm$ .013 & .560 $\pm$ .049 & .880 $\pm$ .021 & .748 $\pm$ .004 \\
    DensMAP & .360 $\pm$ .018 & .596 $\pm$ .010 & .257 $\pm$ .060 & .809 $\pm$ .009 & .510 $\pm$ .043 & .852 $\pm$ .016 & .735 $\pm$ .005 \\
    DensMAP-800 & .372 $\pm$ .022 & .588 $\pm$ .004 & .241 $\pm$ .106 & .809 $\pm$ .024 & .527 $\pm$ .039 & .875 $\pm$ .020 & .741 $\pm$ .008 \\
    PaCMAP & .301 $\pm$ .053 & \underline{.613 $\pm$ .001} & \textbf{.392 $\pm$ .081} & \textbf{.890 $\pm$ .005} & \textbf{.678 $\pm$ .004} & \textbf{.936 $\pm$ .003} & .779 $\pm$ .012 \\
    Scale-PaCMAP & \textbf{.416 $\pm$ .014} & .600 $\pm$ .006 & .335 $\pm$ .050 & .877 $\pm$ .017 & .568 $\pm$ .003 & .876 $\pm$ .059 & .954 $\pm$ .003 \\
    \bottomrule
  \end{tabular}
  }
\end{table}

\begin{table}[htbp]
  \caption{Random triplet}
  \label{tab:supplement-metrics-random-triplet}
  \centering
  \small
  \setlength{\tabcolsep}{3pt}
  \makebox[\textwidth][c]{%
  \begin{tabular}{lccccccc}
    \toprule
    Embedding & MNIST & FMNIST & COIL-20 & Mammoth & Transcriptomics & HSI & Flow cytometry \\
    \midrule
    ScaleMAP & .629 $\pm$ .007 & .718 $\pm$ .004 & .607 $\pm$ .026 & .813 $\pm$ .004 & .700 $\pm$ .009 & .864 $\pm$ .017 & \textbf{.919 $\pm$ .000} \\
    ScaleMAP-200 & .629 $\pm$ .007 & \underline{.733 $\pm$ .004} & .600 $\pm$ .031 & .804 $\pm$ .007 & .692 $\pm$ .011 & .842 $\pm$ .001 & .897 $\pm$ .003 \\
    UMAP & .609 $\pm$ .003 & .728 $\pm$ .003 & .595 $\pm$ .025 & .807 $\pm$ .014 & .685 $\pm$ .004 & .822 $\pm$ .011 & .785 $\pm$ .004 \\
    UMAP-800 & .610 $\pm$ .006 & .719 $\pm$ .002 & .592 $\pm$ .045 & .811 $\pm$ .005 & .692 $\pm$ .018 & .850 $\pm$ .026 & .792 $\pm$ .003 \\
    DensMAP & .623 $\pm$ .004 & .731 $\pm$ .002 & .597 $\pm$ .018 & .805 $\pm$ .007 & .680 $\pm$ .015 & .828 $\pm$ .026 & .782 $\pm$ .003 \\
    DensMAP-800 & .624 $\pm$ .007 & .729 $\pm$ .002 & .590 $\pm$ .033 & .805 $\pm$ .010 & .683 $\pm$ .011 & .829 $\pm$ .028 & .788 $\pm$ .005 \\
    PaCMAP & .613 $\pm$ .013 & \textbf{.734 $\pm$ .001} & \textbf{.662 $\pm$ .030} & \textbf{.874 $\pm$ .003} & \textbf{.738 $\pm$ .001} & \textbf{.885 $\pm$ .003} & .839 $\pm$ .003 \\
    Scale-PaCMAP & \textbf{.637 $\pm$ .006} & .725 $\pm$ .002 & .637 $\pm$ .024 & .854 $\pm$ .008 & .707 $\pm$ .002 & .841 $\pm$ .026 & .898 $\pm$ .003 \\
    \bottomrule
  \end{tabular}
  }
\end{table}

\begin{table}[htbp]
  \caption{Runtime (s)}
  \label{tab:supplement-metrics-runtime}
  \centering
  \small
  \setlength{\tabcolsep}{3pt}
  \makebox[\textwidth][c]{%
  \begin{tabular}{lccccccc}
    \toprule
    Embedding & MNIST & FMNIST & COIL-20 & Mammoth & Transcriptomics & HSI & Flow cytometry \\
    \midrule
    ScaleMAP & 23.9 $\pm$ 0.9 & 27.7 $\pm$ 0.6 & 2.0 $\pm$ 0.0 & 3.3 $\pm$ 0.2 & 131.7 $\pm$ 66.4 & 151.3 $\pm$ 2.8 & 598.2 $\pm$ 53.5 \\
    ScaleMAP-200 & 12.0 $\pm$ 0.2 & 14.9 $\pm$ 0.5 & 1.8 $\pm$ 0.0 & \textbf{2.1 $\pm$ 0.1} & 79.0 $\pm$ 6.7 & \underline{101.4 $\pm$ 2.9} & 512.9 $\pm$ 150.7 \\
    UMAP & \textbf{8.8 $\pm$ 0.4} & \textbf{11.2 $\pm$ 0.6} & 1.4 $\pm$ 0.0 & 2.4 $\pm$ 0.2 & \textbf{69.9 $\pm$ 4.0} & \textbf{97.4 $\pm$ 11.8} & 421.7 $\pm$ 90.8 \\
    UMAP-800 & 18.3 $\pm$ 0.6 & 21.2 $\pm$ 0.6 & 1.5 $\pm$ 0.0 & 2.9 $\pm$ 0.2 & 116.9 $\pm$ 71.2 & 141.5 $\pm$ 6.2 & 490.3 $\pm$ 100.1 \\
    DensMAP & 25.0 $\pm$ 0.7 & 29.0 $\pm$ 1.0 & 2.5 $\pm$ 0.2 & 4.9 $\pm$ 0.2 & 136.3 $\pm$ 45.2 & 175.0 $\pm$ 8.4 & 567.4 $\pm$ 162.5 \\
    DensMAP-800 & 32.5 $\pm$ 0.8 & 36.9 $\pm$ 1.3 & 2.5 $\pm$ 0.1 & 5.1 $\pm$ 0.2 & 165.4 $\pm$ 66.4 & 209.3 $\pm$ 14.6 & 635.4 $\pm$ 148.6 \\
    PaCMAP & 37.2 $\pm$ 0.8 & 36.1 $\pm$ 1.1 & \textbf{0.6 $\pm$ 0.0} & 3.9 $\pm$ 0.1 & 103.5 $\pm$ 2.9 & \underline{206.4 $\pm$ 255.7} & \textbf{216.4 $\pm$ 15.7} \\
    Scale-PaCMAP & 89.7 $\pm$ 1.1 & 88.5 $\pm$ 0.9 & 1.5 $\pm$ 0.0 & 10.6 $\pm$ 0.3 & 223.0 $\pm$ 6.6 & 178.3 $\pm$ 7.4 & 489.2 $\pm$ 80.0 \\
    \bottomrule
  \end{tabular}
  }
\end{table}

\clearpage
\subsection{Additional figures}

\begin{figure}[h]
\centering
\includegraphics[width=1.0\textwidth]{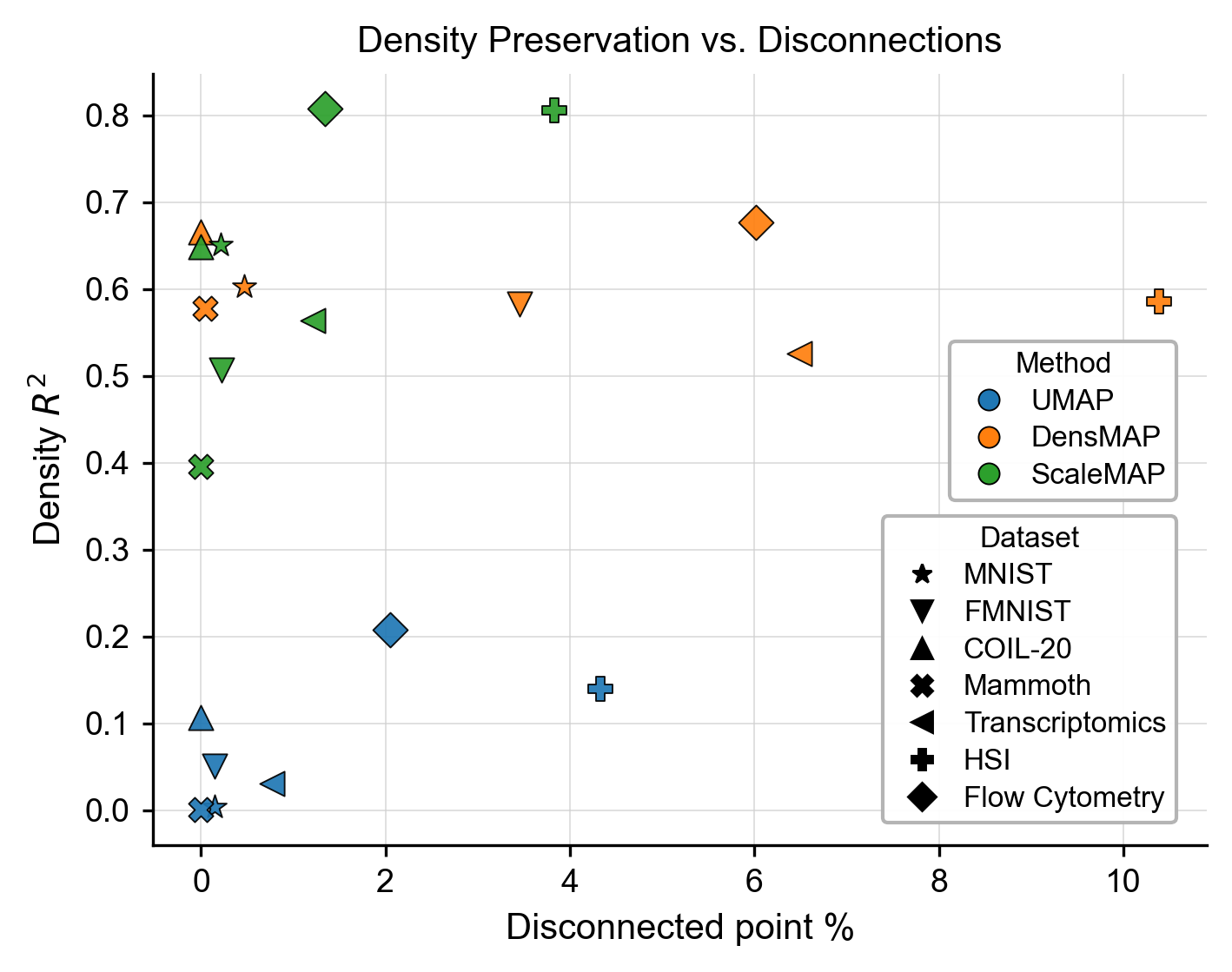}
\caption{\textbf{Tradeoff frontier.} Neighborhood preservation (x-axis, misplaced-point fraction, lower is better) versus density preservation (y-axis, $R^2$, higher is better). Desirable quadrant is top left. Each marker corresponds to a (dataset, method, seed) combination; markers distinguished by method, color by dataset family. ScaleMAP broadly accomplishes both with overall very little tradeoff.}
\label{fig:tradeoff}
\end{figure}

\begin{figure}[p]
\centering
\includegraphics[angle=270,width=0.3\textheight]{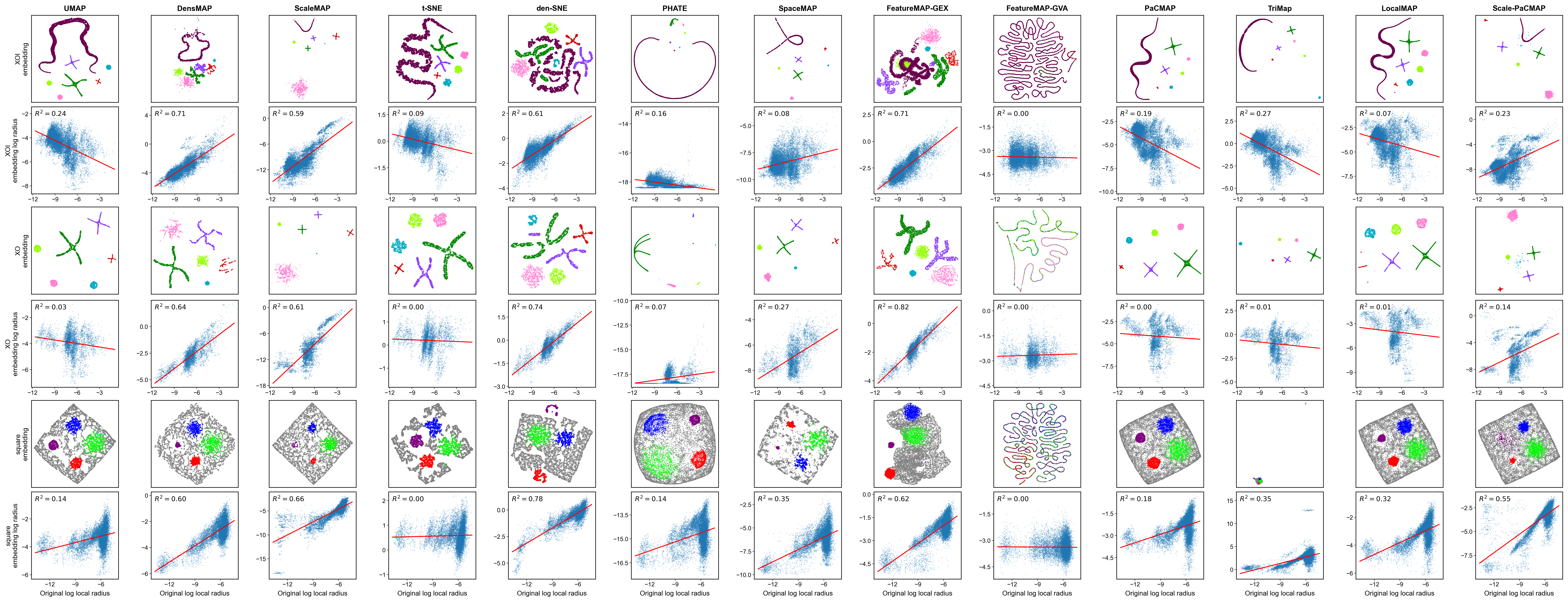}
\caption{\textbf{Synthetic density-conservation benchmark across dimensionality-reduction methods.}
Embeddings and density-preservation fits are shown for the XOI, XO, and square synthetic datasets. Columns correspond to UMAP, DensMAP, ScaleMAP, t-SNE, den-SNE, PHATE, SpaceMAP, FeatureMAP-GEX, FeatureMAP-GVA, PaCMAP, TriMap, LocalMAP, and Scale-PaCMAP. For each dataset, the embedding panel is paired with a density-preservation panel comparing original-space log local radius with embedding-space log local radius; red lines show linear fits and reported $R^2$ values quantify density conservation. Among these, the ones with favorable density preservation are DensMAP, ScaleMAP, den-SNE, FeatureMAP-GEX, and Scale-PaCMAP. Den-SNE could not embed MNIST, one of the medium sized datasets, in 1 hour and was hence excluded. FeatureMAP-GEX, although its density $R^2$ was comparable to DensMAP, generally showed poor embedding quality. For instance, in MNIST, its recall@15 was 4.5\% vs DensMAP's 7.0\% and UMAP's 11.9\%, and in F-MNIST, the respective scores were 4.3\%, 7.1\% and 13.3\%. It also ran out of memory on our machine in the Transcriptomics dataset, and was hence excluded.}
\label{fig:all_dr_comparison}
\end{figure}

\begin{figure}[p]
\centering
\includegraphics[width=1.0\textwidth]{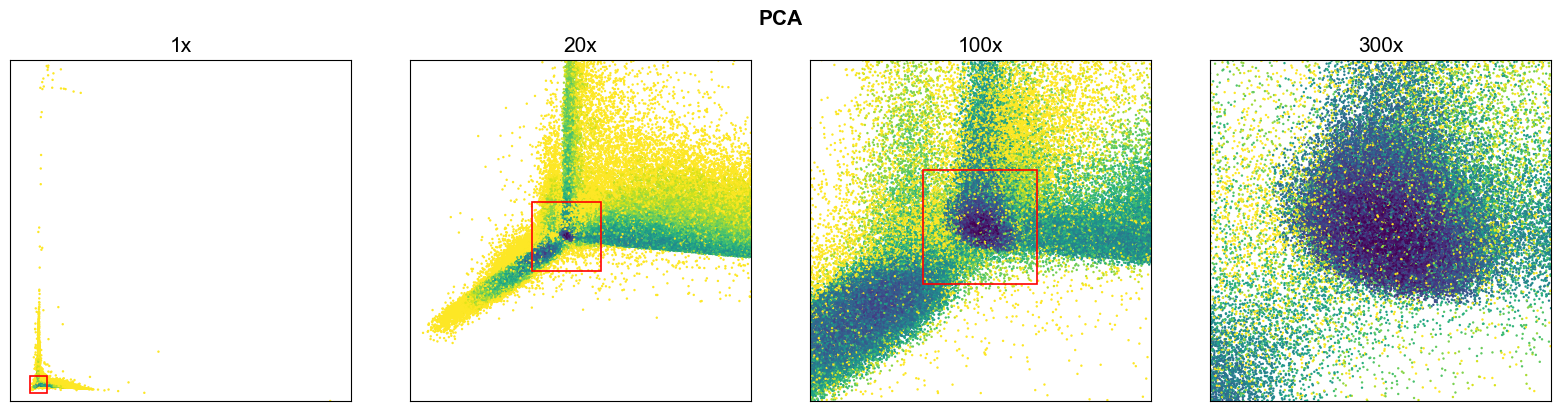}
\caption{\textbf{PCA of Flow cytometry dataset} Coloring matches Figure \ref{fig:flow}. PCA reflects the multiscale structure from Figure \ref{fig:flow}, including a dense cluster of cells that requires significant zoom to observe.}
\label{fig:flow_pca}
\end{figure}

\begin{figure}[p]
\centering
\includegraphics[width=1.0\textwidth]{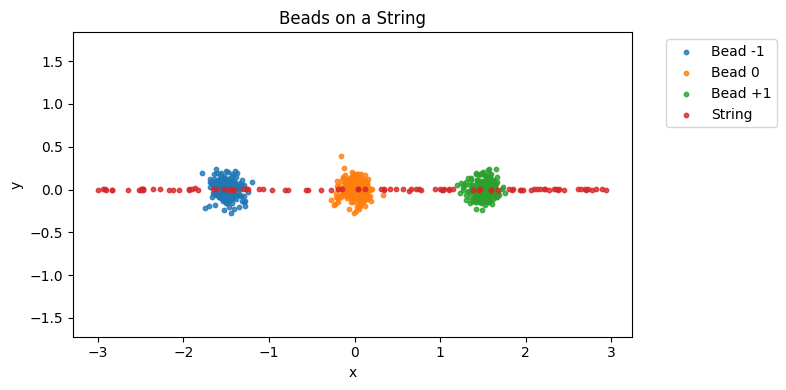}
\caption{\textbf{Beads on string dataset} Three Gaussians connected by a thin uniform distribution.}
\label{fig:beads_on_string}
\end{figure}

\begin{figure}[p]
\centering
\includegraphics[angle=270,width=0.45\textheight]{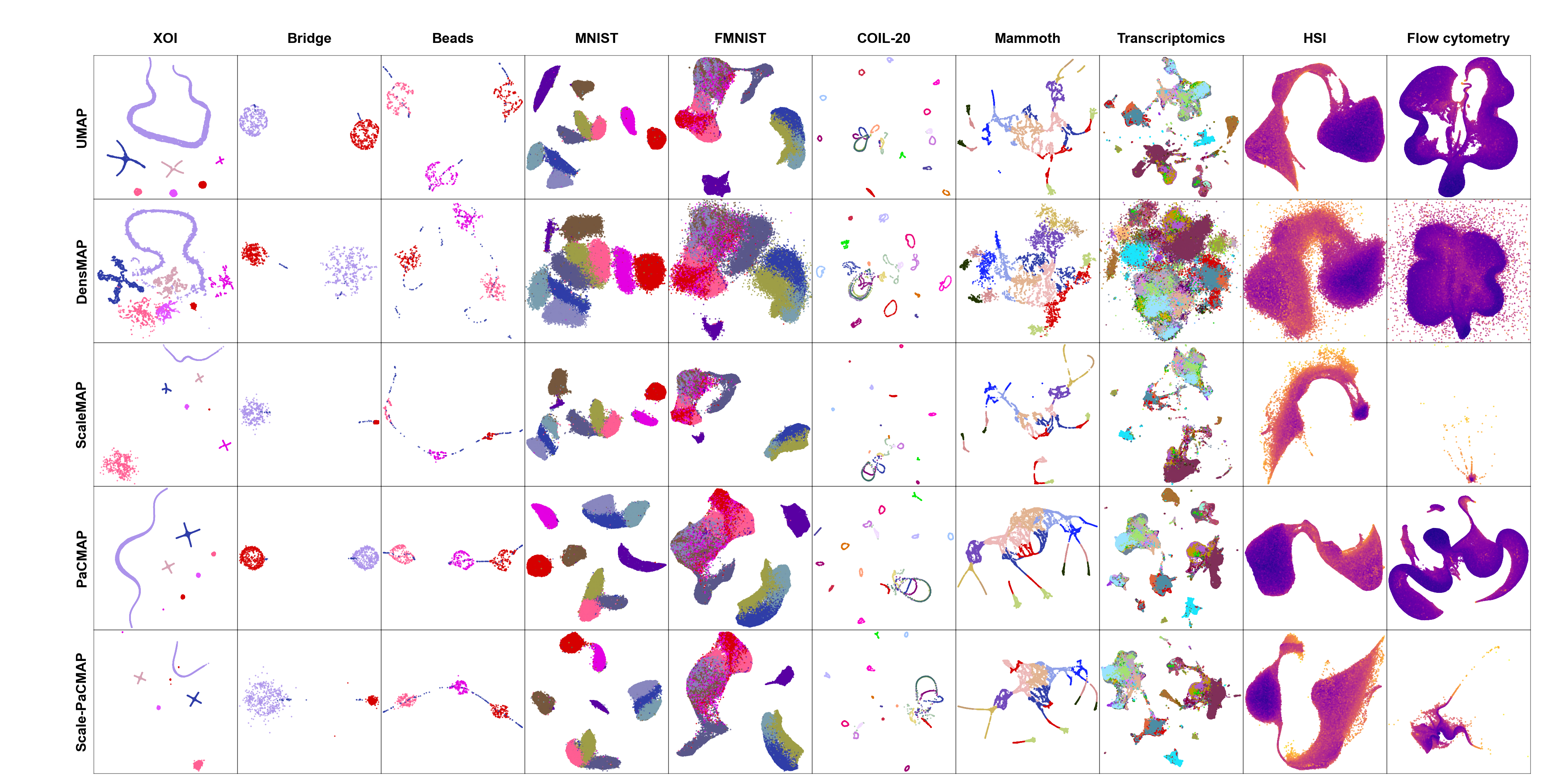}
\caption{\textbf{All datasets vs embedding methods.}
Including PaCMAP and Scale-PaCMAP. Scale-PaCMAP does well in general, except for a tendency to disconnect some points that needn't be disconnected.}
\label{fig:all_dr_gallery}
\end{figure}

\clearpage

\subsection{Dependence of embedding quality on local radius normalization percentile}
\label{sec:percentile_dependence}

We choose $P_{95}$ as a good default for normalizing the local radius for embedding because it works well across datasets. It appears that embedding quality in 2-D datasets depends strongly on this choice, but this effect does not appear in higher dimensional datasets. 

XOI, unlike most other datasets, is sensitive to the percentile we use for local radius normalization. Here, $P_{50}$ gives us very high $R^2$ but does not preserve global structure well. $P_{98}$ gives us very high global structure preservation but poor $R^2$. $P_{95}$ offers a good balance. 

Below, we show the dependence of the ScaleMAP embedding quality on the choice of percentile for a few different datasets.

\begin{figure}[h]
\centering
\includegraphics[width=1.0\textwidth]{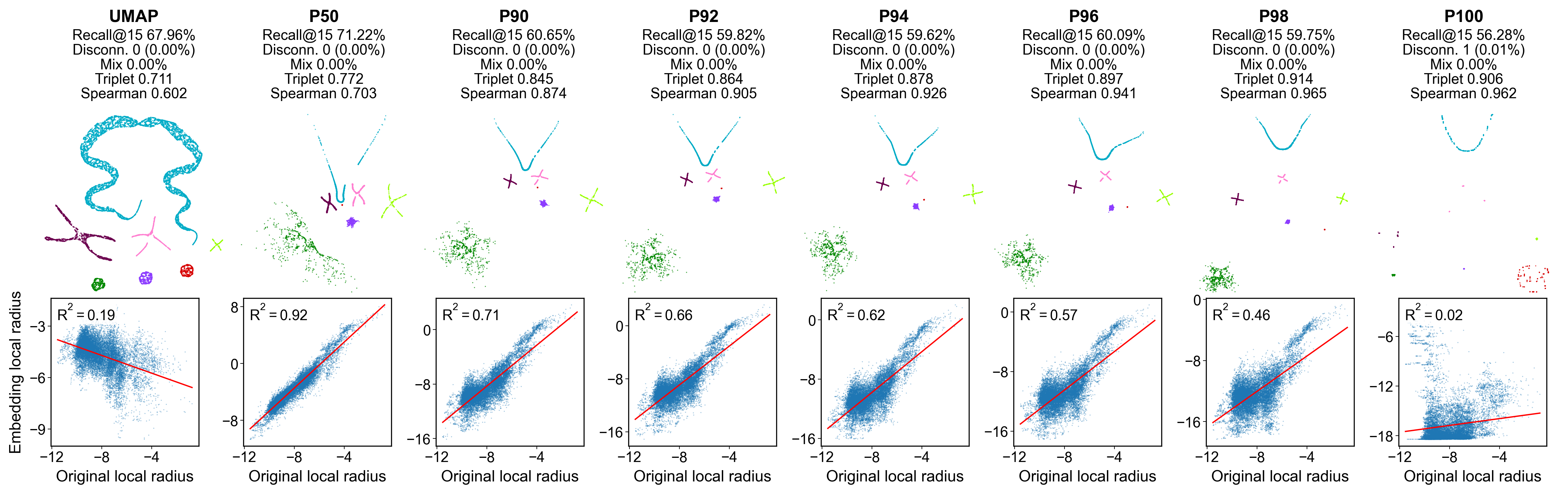}
\caption{\textbf{XOI}}
\label{fig:p95_xoi}
\end{figure}

\begin{figure}[h]
\centering
\includegraphics[width=1.0\textwidth]{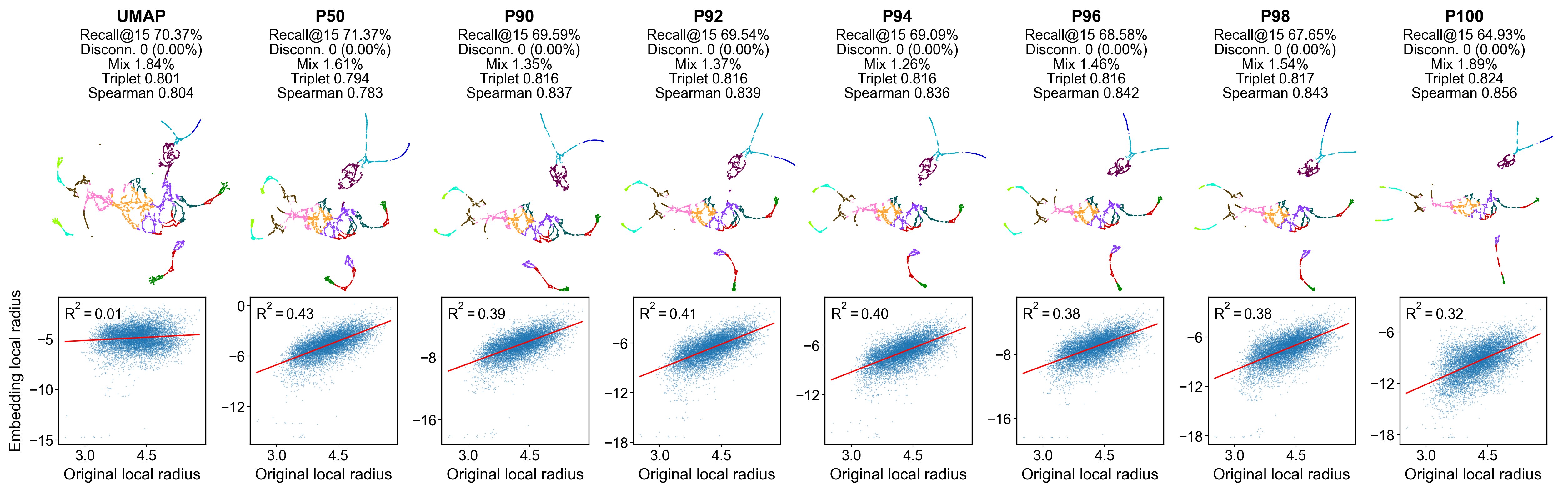}
\caption{\textbf{Mammoth}}
\label{fig:p95_mammoth}
\end{figure}

\begin{figure}[h]
\centering
\includegraphics[width=1.0\textwidth]{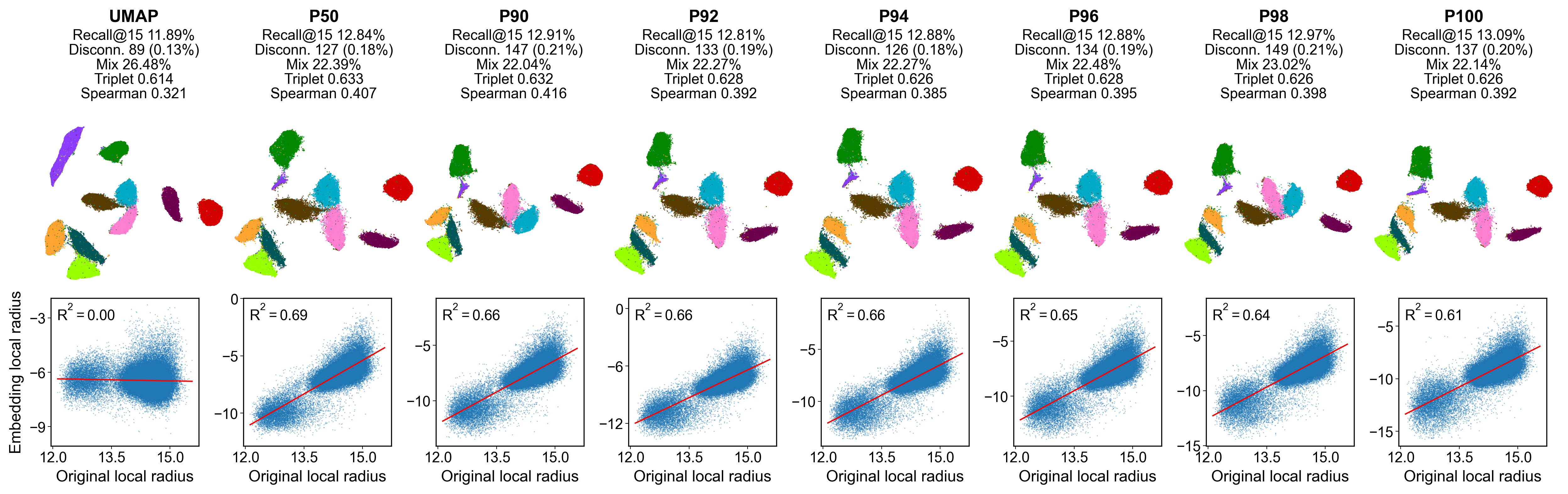}
\caption{\textbf{MNIST}}
\label{fig:p95_mnist}
\end{figure}

\begin{figure}[ht]
\centering
\includegraphics[width=1.0\textwidth]{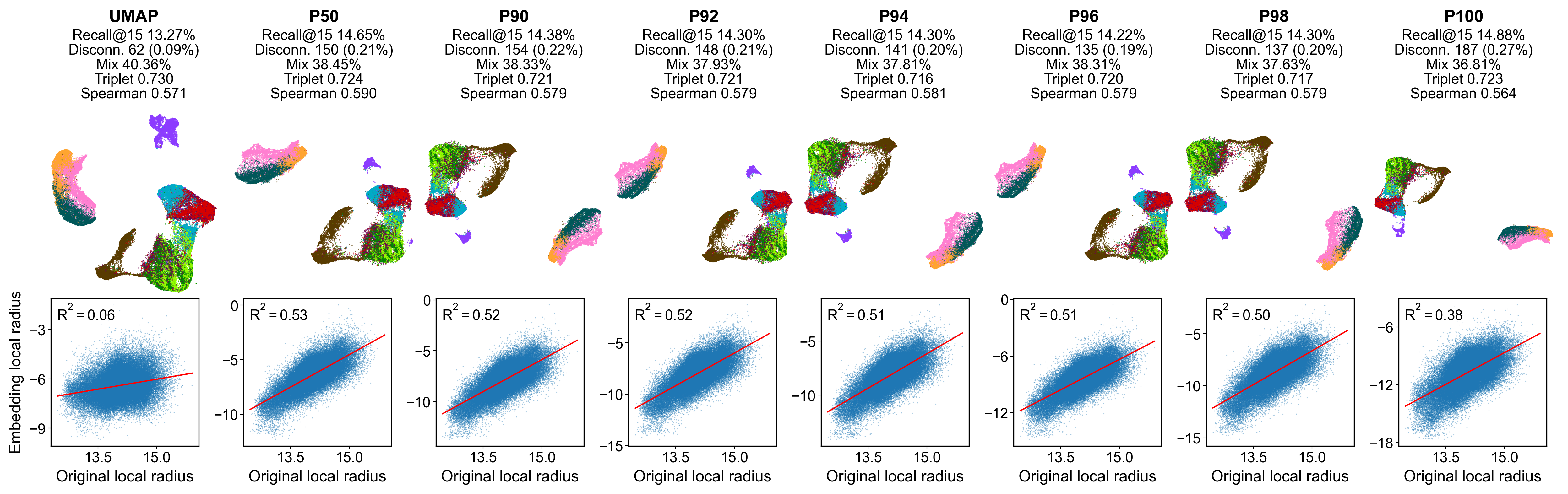}
\caption{\textbf{F-MNIST}}
\label{fig:p95_fmnist}
\end{figure}

\clearpage


\begin{thebibliography}{99}

\bibitem{mcinnes2018umap}
L.~McInnes, J.~Healy, and J.~Melville.
\newblock UMAP: Uniform manifold approximation and projection for dimension reduction.
\newblock \emph{arXiv preprint arXiv:1802.03426}, 2018.

\bibitem{maaten2008visualizing}
L.~van~der Maaten and G.~Hinton.
\newblock Visualizing data using t-SNE.
\newblock \emph{Journal of Machine Learning Research}, 9:2579--2605, 2008.

\bibitem[Wang et~al.(2021)]{wang2021understanding}
Y.~Wang, H.~Huang, C.~Rudin, and Y.~Shaposhnik.
\newblock Understanding how dimension reduction tools work: An empirical approach to deciphering t-SNE, UMAP, TriMap, and PaCMAP for data visualization.
\newblock \emph{Journal of Machine Learning Research}, 22(201):1--73, 2021.

\bibitem{narayan2021assessing}
A.~Narayan, B.~Berger, and H.~Cho.
\newblock Assessing single-cell transcriptomic variability through density-preserving data visualization.
\newblock \emph{Nature Biotechnology}, 39:765--774, 2021.
\newblock doi:10.1038/s41587-020-00801-7.

\bibitem{belkin2003laplacian}
M.~Belkin and P.~Niyogi.
\newblock Laplacian eigenmaps for dimensionality reduction and data representation.
\newblock \emph{Neural Computation}, 15(6):1373--1396, 2003.
\newblock doi:10.1162/089976603321780317.

\bibitem{lecunmnist}
Y.~LeCun, C.~Cortes, and C.~J.~C. Burges.
\newblock The MNIST database of handwritten digits.
\newblock Computer Vision Data Foundation mirror.
\newblock \url{https://github.com/cvdfoundation/mnist}.
\newblock Accessed May 7, 2026.

\bibitem{xiao2017fashion}
H.~Xiao, K.~Rasul, and R.~Vollgraf.
\newblock Fashion-MNIST: A novel image dataset for benchmarking machine learning algorithms.
\newblock \emph{arXiv preprint arXiv:1708.07747}, 2017.

\bibitem{nene1996coil20}
S.~A. Nene, S.~K. Nayar, and H.~Murase.
\newblock Columbia Object Image Library (COIL-20).
\newblock Technical Report CUCS-005-96, Department of Computer Science, Columbia University, February 1996.

\bibitem{smithsonianmammoth}
Smithsonian Institution.
\newblock \emph{Mammuthus primigenius} (Blumbach).
\newblock Smithsonian 3D Digitization, National Museum of Natural History, Paleobiology Department, Record ID \texttt{nmnhpaleobiology\_3447777}.
\newblock \url{https://3d.si.edu/object/3d/mammuthus-primigenius-blumbach:341c96cd-f967-4540-8ed1-d3fc56d31f12}.
\newblock Accessed May 7, 2026.

\bibitem{tabulasapiens2022}
Tabula Sapiens Consortium.
\newblock The Tabula Sapiens: A multiple-organ, single-cell transcriptomic atlas of humans.
\newblock \emph{Science}, 376(6594):eabl4896, 2022.
\newblock doi:10.1126/science.abl4896.

\bibitem{lopez2018deep}
R.~Lopez, J.~Regier, M.~B.~Cole, M.~I.~Jordan, and N.~Yosef.
\newblock Deep generative modeling for single-cell transcriptomics.
\newblock \emph{Nature Methods}, 15(12):1053--1058, 2018.
\newblock doi:10.1038/s41592-018-0229-2.

\bibitem{camp2014highspeed}
C.~H. Camp Jr., Y.~J. Lee, J.~M. Heddleston, C.~M. Hartshorn, A.~R. Hight Walker, J.~N. Rich, J.~D. Lathia, and M.~T. Cicerone.
\newblock High-speed coherent Raman fingerprint imaging of biological tissues.
\newblock \emph{Nature Photonics}, 8:627--634, 2014.
\newblock doi:10.1038/nphoton.2014.145.

\bibitem{poorna2023toward}
R.~Poorna, W.-W. Chen, A.~Germond, P.~Qiu, and M.~T. Cicerone.
\newblock Toward gene-correlated spatially resolved metabolomics with fingerprint coherent Raman imaging.
\newblock \emph{The Journal of Physical Chemistry B}, 127(25):5576--5587, 2023.
\newblock doi:10.1021/acs.jpcb.3c01446.

\bibitem{qiu2011spade}
P.~Qiu, E.~F. Simonds, S.~C. Bendall, K.~D. Gibbs Jr., R.~V. Bruggner, M.~D. Linderman, K.~Sachs, G.~P. Nolan, and S.~K. Plevritis.
\newblock Extracting a cellular hierarchy from high-dimensional cytometry data with SPADE.
\newblock \emph{Nature Biotechnology}, 29(10):886--891, 2011.
\newblock doi:10.1038/nbt.1991.

\end{thebibliography}
\end{document}